\documentclass{article}

\PassOptionsToPackage{numbers, compress}{natbib}

\usepackage[preprint]{neurips_2026}
\usepackage[most]{tcolorbox}

\usepackage{amsmath}
\usepackage{amssymb}
\usepackage{listings}
\usepackage{xcolor}
\usepackage{comment}
\usepackage{url}
\usepackage{subcaption}
\usepackage{diagbox}
\usepackage{float}
\usepackage{caption}
\usepackage{pgf}
\usepackage{booktabs}
\usepackage{adjustbox}
\usepackage[colorlinks=true,linkcolor=black,citecolor=black,urlcolor=blue,filecolor=blue]{hyperref}
\usepackage{graphicx} 
\usepackage{tikz}
\usetikzlibrary{positioning, calc, decorations.pathreplacing, arrows.meta}

\definecolor{codegreen}{rgb}{0,0.6,0}
\definecolor{codegray}{rgb}{0.5,0.5,0.5}
\definecolor{codepurple}{rgb}{0.58,0,0.82}
\definecolor{backcolour}{rgb}{0.95,0.95,0.92}

\lstdefinestyle{python}{
    backgroundcolor=\color{backcolour},   
    commentstyle=\color{codegreen},
    keywordstyle=\color{magenta},
    numberstyle=\tiny\color{codegray},
    stringstyle=\color{codepurple},
    basicstyle=\ttfamily\footnotesize,
    breakatwhitespace=false,         
    breaklines=true,                 
    captionpos=b,                    
    keepspaces=true,                 
    numbers=left,                    
    numbersep=5pt,                  
    showspaces=false,                
    showstringspaces=false,
    showtabs=false,                  
    tabsize=2
}

\newtcolorbox{perspective}{
    colback=blue!5!white,
    colframe=blue!75!black,
    fonttitle=\bfseries,
    title=Perspective,
    arc=4pt 
}

\title{Efficient Pre-Training with Token Superposition}
\author{%
  Bowen~Peng\textsuperscript{\textasteriskcentered} \\
  Nous~Research \\
  \texttt{bloc@nousresearch.com} \\
  \And
  Théo~Gigant\textsuperscript{\textasteriskcentered} \\
  Nous~Research \\
  \texttt{theo@nousresearch.com} \\
  \And
  Jeffrey~Quesnelle \\
  Nous~Research \\
  \texttt{emozilla@nousresearch.com} \\
}

\date{May 2026}

\begin{document}

\maketitle

\renewcommand{\thefootnote}{\fnsymbol{footnote}}
\footnotetext[1]{Equal contribution.}
\renewcommand{\thefootnote}{\arabic{footnote}}

\newcommand{\stoken}{"s-token"}
\newcommand{\stokens}{"s-tokens"}

\begin{abstract}
   Pre-training of Large Language Models is often prohibitively expensive and inefficient at scale, requiring complex and invasive modifications in order to achieve high data throughput.
   In this work, we present Token-Superposition Training (TST), a simple drop-in method that significantly improves the data throughput per FLOPs during pre-training without modifying the parallelism, optimizer, tokenizer, data, or model architecture.
   TST is done in two phases: (i) A highly efficient superposition phase where we combine many contiguous tokens into one bag and train using a multi-hot cross-entropy (MCE) objective, and (ii) a recovery phase where we revert back to standard training.
   We extensively evaluate TST on the scale of 270M and 600M parameters and validate on 3B and a 10B A1B mixture of experts model, demonstrating that it is highly robust in different settings. Ultimately, TST consistently outperforms baseline loss and downstream evaluations, and under equal-loss settings, TST yields up to a 2.5x reduction in total pre-training time at the 10B A1B scale.
\end{abstract}

\begin{figure}[h!]
    \centering
    \resizebox{0.6\linewidth}{!}{\input{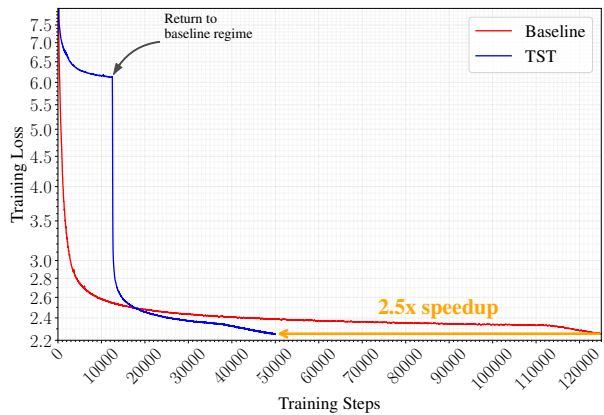}}
    \caption{Loss curves during the pre-training of two Qwen3-like MoE models (10B-A1B) with baseline pretraining and token superposition training (TST) where we stop the training early to match an equal training loss. The baseline training sees 1.05T tokens while the TST training sees 2T data tokens. Every step in all conditions are equal-FLOPs, thus the speedup can be directly computed w.r.t. the number of steps. More details are in Table~\ref{tab:results}.}
    \label{fig:superposition-qwen}
\end{figure}

\section{Introduction}

The rapid proliferation of modern generalist Large Language Models (LLMs) has been driven not only by increases in model size, but critically, by aggressive data scaling~\citep{touvron_llama_2023, grattafiori_llama_2024, bi_deepseek_2024, liu_deepseek-v3_2024, bai_qwen_2023, yang_qwen3_2025, agarwal_gpt-oss-120b_2025}, with recent training regimes often overtraining~\cite{gadre_language_2024} well beyond compute-optimal estimates to maximize performance at inference time.
In this data-hungry paradigm, one of the major concerns during pre-training is the efficiency in which raw text is consumed given a fixed amount of compute.

Recent advances in language modeling pre-training efficiency can be broadly organized into three categories:
\begin{enumerate}
    \item \textbf{Information maximization}: maximizing the information given per sample through improved input priors and representations (better tokenization, BPE~\citep{sennrich_neural_2016}, SuperBPE~\citep{liu_superbpe_2025}, Unigram~\citep{kudo_sentencepiece_2018}, $n$-gram hashing~\citep{liu_scaling_2026, cheng_conditional_2026}) and richer training signals (auxiliary losses, including multi-token prediction~\citep{gloeckle_better_2024, liu_deepseek-v3_2024}, order-augmented objectives~\citep{zuhri_predicting_2026}).
    
    \item \textbf{Compute sparsity}: keeping the input representation fixed but reducing the FLOPs required to process each token by activating only a subset of parameters or attending to a subset of positions. This constitutes a \textit{compute-level} prior: similar expressivity, less work per token (sparse mixture-of-experts \citep{jiang_mixtral_2024, dai_deepseekmoe_2024}, sparse attention \citep{zaheer_big_2020, yuan_native_2025}).

    \item \textbf{Compressive modeling}: learning to \textit{further compress} the representation within the model itself, reducing the number of representations that flow through the expensive layers. This constitutes a \textit{representation-level} prior: fewer tokens internally, but dense computation on each (e.g., Bolmo \citep{minixhofer_bolmo_2025}, H-Net \citep{hwang_dynamic_2025}, Byte-Latent Transformer \citep{pagnoni_byte_2025}, Autoregressive UNet \citep{videau_bytes_2025} Perceiver-Resampler \citep{alayrac_flamingo_2022}).
\end{enumerate}

Note that some of these works also touch on inference-time efficiency, which we acknowledge as an important line of work but orthogonal to our focus on pre-training efficiency.
While some methods above confound and mix both training-time and inference-time efficiency, other methods focus on inference-time efficiency only (\textit{e.g.} diffusion~\citep{nie_large_2025}, speculative decoding~\citep{leviathan_fast_2023}), and almost none of them have decoupled the training-time efficiency and focus solely on this latter part (except concurrent work from~\citet{zheng_proxy_2026}).

Many recent works refute the training-time and inference-time efficiency equivalence, where they show that scaling up inference-time compute independently of training time compute can improve performance on downstream tasks (e.g. CoT/reasoning models~\citep{wei_chain--thought_2022}, ParScale~\citep{chen_parallel_2025}, looped language models~\citep{zhu_scaling_2025}), thus it is important for our method to be only used during training, and keep the model architecture and expressivity ``untouched" for inference compared to the baseline in order to minimize any confounding factors. 

Furthermore, recent evidence suggests that the performance advantage of subword-based over byte-level language models is largely driven by increased training sample throughput \citep{gigant_decoupling_2026}.
Building on this insight, we investigate whether LLM training efficiency can be further optimized by maximizing training-time throughput independently of the model's inference-time architecture.

Finally, the monolithic pre-training paradigm is currently shifting towards two-stage or multi-stage pre-training schemes.
Many recent works have found that better or more efficient training methods can be used for the first stage of training~\citep{minixhofer_bolmo_2025, hu_between_2025, lee_training_2026}, and then find that the model is elastic enough to quickly adapt to the final desired model behavior, often saving on training cost and/or resulting in a better quality model.
However, most prior works focus on the mid-training or post-training regime, but we contextualize this work by showing that the same ideas can be applied to pre-training.

We introduce Token-Superposition Training (TST), a method that tries to improve training efficiency by increasing token throughput while still priming the model for the desired task of autoregressive prediction.
Our work starts with this fundamentally different perspective:

\begin{perspective}
Can we improve pre-training efficiency by forcing a higher token throughput during training, without modifying the final model architecture and its inference dynamics?
\end{perspective}

After completion of this work, we became aware of \citet{shao2025beyond}, which independently proposed the same core mechanism: averaging consecutive token embeddings as input, predicting all tokens in the next group via cross-entropy with a single head, and transferring to standard token-level training in a second phase, with only minor differences from ours on the algorithm side. Although our work converges to a very similar method, we propose a different view on the problem and extend the research in several directions, with more details in Section~\ref{sec:patchlevel}.

\section{Related works}

\subsection{Alternative Prediction Objectives}
In contrast to standard autoregressive next-token prediction, several studies have explored alternative training objectives to improve representation learning. 
\citet{tay_ul2_2022} introduced the \textit{mixture-of-denoisers} framework, which unifies diverse denoising tasks—such as span corruption and causal language modeling to provide superior generalization across architectures.
In the context of encoder models, \citet{gisserot-boukhlef_should_2025} demonstrated that a two-stage pretraining schedule, transitioning from causal language modeling to masked language modeling (MLM), outperforms standard MLM baselines in some downstream tasks.

\subsection{Auxiliary Prediction of Future Representations}

Recent pre-training literature introduced auxiliary losses to move beyond single-token targets, with the goal of increasing the information density per gradient step. 
\citet{gloeckle_better_2024} introduced \textit{multi-token prediction} (MTP), using $k$ independent heads to predict the next $k$ tokens simultaneously.
Although MTP improves sample efficiency in some cases and has been featured in major state-of-the-art LLM pre-training runs \cite{liu_deepseek-v3_2024}, it has limited benefit to smaller models and requires the tuning of an extra hyper-parameter $k$ while introducing additional parameters.

Other approaches explore using auxiliary losses with targets using representations involving future tokens.
DeepseekV3 \cite{liu_deepseek-v3_2024} uses a modified version of MTP, featuring cascaded predictions using additional MTP modules.
\citet{zuhri_predicting_2026} replace MTP with the prediction of the relative order of future tokens, relaxing the complexity of MTP by simplifying the task and requiring only a single additional head. 
\citet{liu_next_2026} proposed \textit{next concept prediction} with predicted concepts covering segments spanning multiple tokens.
\citet{mahajan_beyond_2025} introduced \textit{future summary prediction}, where the model predicts a compressed representation of future tokens utilizing either hand-crafted bag-of-words representations or learned latent features from a reversed sequence model.
Notably, an entry within the \texttt{modded-nanogpt} speedrun \citep{kellerjordan_new_nodate} has implemented an MTP-inspired loss that shares conceptual similarities with the \textit{next bag-of-tokens} prediction explored in our work.
These works illustrate that additional signal involving predicting representations of future tokens can yield substantial gains in pre-training sample efficiency.

\subsection{Input Granularity}

Input granularity is a fundamental hyperparameter in both vision and language modeling.
In Vision Transformers \citep{dosovitskiy_image_2020}, patch size serves as a primary control for the trade-off between FLOPs and performance.
\citet{anagnostidis_navigating_2024} showed that scheduling patch size from coarse to fine-grained during training improves isoFLOPs performance for Vision Transformers.
In LLMs, this granularity is typically set by the tokenizer's \textit{fertility}, \textit{i.e.} the average number of tokens required to represent a word.
Subword tokenizers provide a coarser-grained view of text compared to byte-level or unicode representations.
\citet{liu_superbpe_2025} further merge BPE tokens into supertokens, resulting in coarser tokens exhibiting improved training performance compared to regular BPE.
Recent work by \citet{minixhofer_bolmo_2025} explored coarse-to-fine distillation of LLMs by resuming subword-level pretraining with byte-level objectives.
Furthermore, \citet{zheng_proxy_2026} demonstrated that including mixed granularities at training-time, through both compressed and raw byte representations, resulted in better byte-level performance compared to raw byte representation only.
\citet{gigant_decoupling_2026} investigated the performance gap between subword and byte-level models, attributing the subword advantage largely to increased \textit{sample throughput} at isoFLOPs, a direct consequence of the coarser tokens resulting from the sequence compression effect of subword tokenization.
These findings suggest that coarser training-time input granularity is a key driver of modern LLM training speed.
In this work, we leverage both the additional signal from future tokens representations and a coarser input granularity at training-time to improve the model pre-training efficiency. Crucially, during the second phase of training, we return to the baseline granularity and loss.

\subsection{Patch-Level Training}
\label{sec:patchlevel}
Most directly related to our work, \citet{shao2025beyond} introduced patch-level training for LLMs, in which consecutive token embeddings are averaged into "patches" and the model trained to predict all tokens in the next patch using a single output head with cross-entropy loss. After patch-level training, the model reverts to standard token-level training. This is algorithmically identical to what we term Token Superposition Training, but with major differences in background theory and execution, which we attribute to conceptual convergence. \citet{shao2025beyond} frame patch-level training as reducing total FLOPs for a fixed dataset, whereas we propose TST as two orthogonal but additive processes that increase token throughput at constant per-step FLOPs (Section~\ref{sec:inputoutput}) following the throughput hypothesis of~\citet{gigant_decoupling_2026}, where we make careful implementation decisions that result in accurate comparisons which simplifies scaling, and justify the choice of the multi-hot loss function based on empirical evidence (Section~\ref{sec:outputsuperposition}). \citet{shao2025beyond} demonstrated the approach with trainings up to 2.7B parameters on 360B tokens, achieving a total training cost of 0.5$\times$. We extend this with an extensive search of the hyperparameter space and demonstrate successful scaling at the 10B parameters, 2T token scale. They also observed that maintaining the same architecture (without additional projection layers) across both phases is critical for successful recovery. This finding aligns with our representation alignment analysis (Section~\ref{sec:twophasealign}).

\section{Methodology}

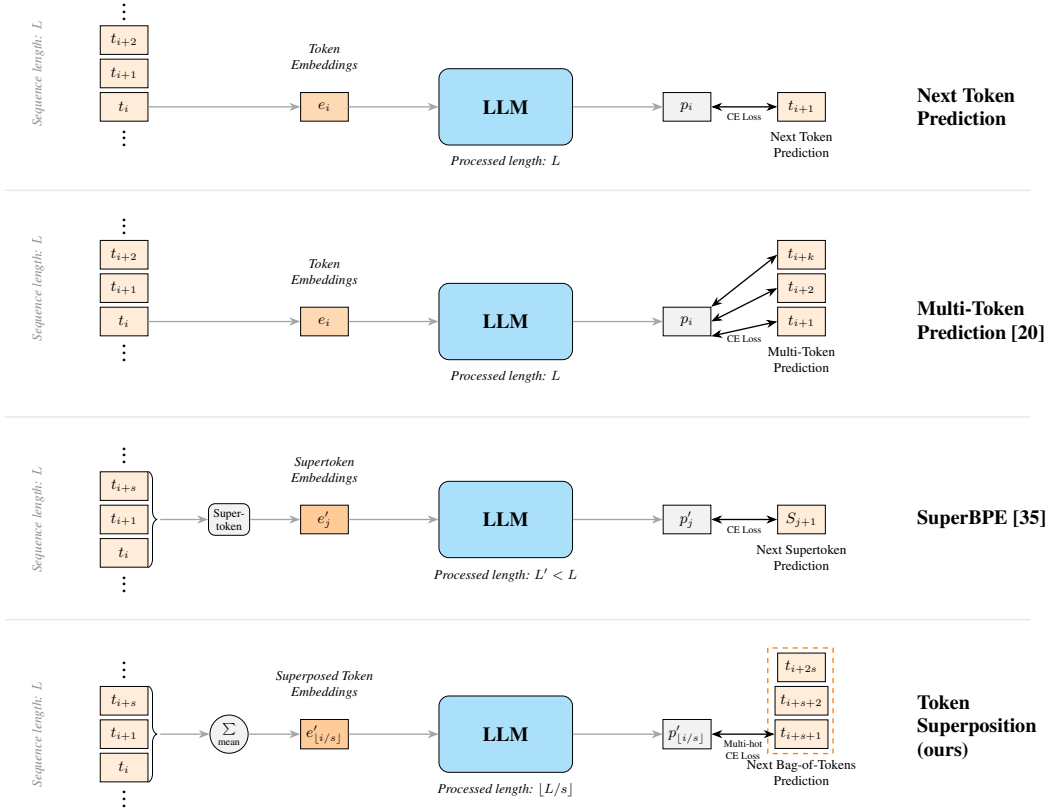
\begin{figure}[bht]
    \centering
    \resizebox{\linewidth}{!}{
    \begin{tikzpicture}[
        node distance=1cm and 1.5cm,
        box/.style={rectangle, draw, fill=orange!15, minimum width=1.0cm, minimum height=0.6cm, font=\small},
        llm/.style={rectangle, draw, fill=cyan!30, rounded corners=6pt, minimum width=2.8cm, minimum height=1.6cm, font=\bfseries\large},
        pred/.style={rectangle, draw, fill=gray!10, minimum width=1.0cm, minimum height=0.6cm, font=\small},
        mean/.style={circle, draw, fill=gray!10, inner sep=0pt, minimum size=0.8cm, align=center, font=\scriptsize},
        supernode/.style={rectangle, rounded corners=4pt, draw, fill=gray!10, inner sep=3pt, align=center, font=\scriptsize},
        arrow/.style={-Stealth, thick, color=gray!70},
        loss_arrow/.style={Stealth-Stealth, thick, color=black},
        label_text/.style={font=\bfseries\large, align=left, anchor=west} 
    ]
    
    \def\colSeqLen{-1.8}
    \def\colTokens{0}
    \def\colMean{2.2}
    \def\colEmbed{4.2}
    \def\colLLM{8.0}
    \def\colPred{11.8}
    \def\colTarget{14.2}
    \def\colRegime{16.5} 
    
    \begin{scope}[name prefix=base-]
        \node[rotate=90, font=\footnotesize\itshape, color=gray!90] at (\colSeqLen, 0.7) {Sequence length: $L$};
    
        \node[font=\Large] at (\colTokens, 2.1) {$\vdots$};
        \node[box] (tip2) at (\colTokens, 1.4) {$t_{i+2}$};
        \node[box] (tip1) at (\colTokens, 0.7) {$t_{i+1}$};
        \node[box] (ti) at (\colTokens, 0) {$t_i$};
        \node[font=\Large] at (\colTokens, -0.55) {$\vdots$};
    
        \node[box, fill=orange!30] (ei) at (\colEmbed, 0) {$e_i$};
        \node[above=0.3cm of ei, font=\footnotesize\itshape, align=center] {Token\\Embeddings};
        \draw[arrow] (ti.east) -- (ei.west);
    
        \node[llm] (llm) at (\colLLM, 0) {LLM};
        \node[below=0.1cm of llm, font=\footnotesize\itshape] {Processed length: $L$};
        \draw[arrow] (ei.east) -- (llm.west);
    
        \node[pred] (pi) at (\colPred, 0) {$p_i$};
        \node[box] (target) at (\colTarget, 0) {$t_{i+1}$};
        \draw[arrow] (llm.east) -- (pi.west);
        \draw[loss_arrow] (pi) -- (target) node[midway, below, font=\tiny] {CE Loss};
        \node[below=0.1cm of target, font=\footnotesize, align=center] {Next Token\\Prediction};
    
        \node[label_text] at (\colRegime, 0) {Next Token\\Prediction};
    \end{scope}
    
    \draw[thick, gray!20] (-2.5, -1.75) -- (\colRegime+2.5, -1.75);
    
    \begin{scope}[yshift=-4.5cm, name prefix=mtp-]
        \node[rotate=90, font=\footnotesize\itshape, color=gray!90] at (\colSeqLen, 0.7) {Sequence length: $L$};
    
        \node[font=\Large] at (\colTokens, 2.1) {$\vdots$};
        \node[box] (tip2) at (\colTokens, 1.4) {$t_{i+2}$};
        \node[box] (tip1) at (\colTokens, 0.7) {$t_{i+1}$};
        \node[box] (ti) at (\colTokens, 0) {$t_i$};
        \node[font=\Large] at (\colTokens, -0.55) {$\vdots$};
    
        \node[box, fill=orange!30] (ei) at (\colEmbed, 0) {$e_i$};
        \node[above=0.3cm of ei, font=\footnotesize\itshape, align=center] {Token\\Embeddings}; 
        \draw[arrow] (ti.east) -- (ei.west);
    
        \node[llm] (llm) at (\colLLM, 0) {LLM};
        \node[below=0.1cm of llm, font=\footnotesize\itshape] {Processed length: $L$};
        \draw[arrow] (ei.east) -- (llm.west);
    
        \node[pred] (pi) at (\colPred, 0) {$p_i$};
        
        \node[box] (b1) at (\colTarget, 0) {$t_{i+1}$};
        \node[box] (b2) at (\colTarget, 0.7) {$t_{i+2}$};
        \node[box] (b3) at (\colTarget, 1.4) {$t_{i+k}$}; 
        
        
        \draw[arrow] (llm.east) -- (pi.west);
        
        \draw[loss_arrow] (pi.south east) -- (b1.west) node[midway, below, font=\tiny] {CE Loss};
        \draw[loss_arrow] (pi.east) -- (b2.west);
        \draw[loss_arrow] (pi.north east) -- (b3.west);
        
        \node[below=0.1cm of b1, font=\footnotesize, align=center] {Multi-Token\\Prediction};
    
        \node[label_text] at (\colRegime, 0) {Multi-Token\\Prediction \citep{gloeckle_better_2024}};
    \end{scope}
    
    \draw[thick, gray!20] (-2.5, -6.5) -- (\colRegime+2.5, -6.5);
    
    \begin{scope}[yshift=-9.0cm, name prefix=bpe-]
        \node[rotate=90, font=\footnotesize\itshape, color=gray!90] at (\colSeqLen, 0.35) {Sequence length: $L$};
    
        \node[font=\Large] at (\colTokens, 1.8) {$\vdots$};
        \node[box] (tip2) at (\colTokens, 1.05) {$t_{i+s}$};
        \node[box] (tip1) at (\colTokens, 0.35) {$t_{i+1}$};
        \node[box] (ti) at (\colTokens, -0.35) {$t_i$};
        \node[font=\Large] at (\colTokens, -0.9) {$\vdots$};
    
        \draw[decorate, decoration={brace, amplitude=5pt}] (tip2.north east) -- (ti.south east);
        
        \node[supernode] (m1) at (\colMean, 0.35) {Super-\\token};
        \node[box, fill=orange!40] (e1) at (\colEmbed, 0.35) {$e'_{j}$};
        \node[above=0.3cm of e1, font=\footnotesize\itshape, align=center] {Supertoken\\Embeddings}; 
        
        \draw[arrow] ($(tip1.east)+(0.25,0)$) -- (m1.west);
        \draw[arrow] (m1.east) -- (e1.west);
    
        \node[llm] (llm) at (\colLLM, 0.35) {LLM};
        \node[below=0.1cm of llm, font=\footnotesize\itshape] {Processed length: $L' < L$};
        \draw[arrow] (e1.east) -- (llm.west);
    
        \node[pred] (pi) at (\colPred, 0.35) {$p'_{j}$};
        \node[box] (b1) at (\colTarget, 0.35) {$S_{j+1}$}; 
        
        \draw[arrow] (llm.east) -- (pi.west);
        \draw[loss_arrow] (pi.east) -- (b1.west) node[midway, below, font=\tiny] {CE Loss};
        \node[below=0.1cm of b1, font=\footnotesize, align=center] {Next Supertoken\\Prediction};
    
        \node[label_text] at (\colRegime, 0.35) {SuperBPE \citep{liu_superbpe_2025}};
    \end{scope}
    
    \draw[thick, gray!20] (-2.5, -10.75) -- (\colRegime+2.5, -10.75);
    
    \begin{scope}[yshift=-13.5cm, name prefix=super-]
        \node[rotate=90, font=\footnotesize\itshape, color=gray!90] at (\colSeqLen, 0.35) {Sequence length: $L$};
    
        \node[font=\Large] at (\colTokens, 1.8) {$\vdots$};
        \node[box] (tip2) at (\colTokens, 1.05) {$t_{i+s}$};
        \node[box] (tip1) at (\colTokens, 0.35) {$t_{i+1}$};
        \node[box] (ti) at (\colTokens, -0.35) {$t_i$};
        \node[font=\Large] at (\colTokens, -0.9) {$\vdots$};
    
        \draw[decorate, decoration={brace, amplitude=5pt}] (tip2.north east) -- (ti.south east);
        \node[mean] (m1) at (\colMean, 0.35) {$\sum$ \\ \tiny mean};
        \node[box, fill=orange!40] (e1) at (\colEmbed, 0.35) {$e'_{\lfloor i/s \rfloor}$}; 
        
        \draw[arrow] ($(tip1.east)+(0.25,0)$) -- (m1.west);
        \draw[arrow] (m1.east) -- (e1.west);
        
        \node[above=0.3cm of e1, font=\footnotesize\itshape, align=center] {Superposed Token\\Embeddings};
    
        \node[llm] (llm) at (\colLLM, 0.35) {LLM};
        \node[below=0.1cm of llm, font=\footnotesize\itshape] {Processed length: $\lfloor L/s \rfloor$}; 
        \draw[arrow] (e1.east) -- (llm.west);
    
        \node[pred] (pi) at (\colPred, 0.35) {$p'_{\lfloor i/s \rfloor}$}; 
        \node[box] (b1) at (\colTarget, 0.35) {$t_{i+s+1}$};
        \node[box] (b2) at (\colTarget, 1.05) {$t_{i+s+2}$};
        \node[box] (b3) at (\colTarget, 1.75) {$t_{i+2s}$};
        
        \draw[dashed, orange, thick] ($(b3.north west)+(-0.2,0.1)$) rectangle ($(b1.south east)+(0.1,-0.1)$);
        
        \draw[arrow] (llm.east) -- (pi.west);
        \draw[loss_arrow] (pi.east) -- (b1.west) node[midway, below, font=\tiny, align=center] {Multi-hot\\CE Loss};
        \node[below=0.1cm of b1, font=\footnotesize, align=center] {Next Bag-of-Tokens\\Prediction};
    
        \node[label_text] at (\colRegime, 0.5) {Token\\Superposition\\(ours)};
    \end{scope}
    \end{tikzpicture}
    
    }
    \caption{Comparison between standard next token prediction, TST and a few methods that superficially resemble TST. Note that this comparison is illustrative only for the purpose of understanding TST.}
    \label{superposition-tikz}
\end{figure}

Token Superposition involves two small changes when compared to baseline next-token prediction training, illustrated in Figure \ref{superposition-tikz}.
In the Token Superposition Training (TST) regime, an LLM processes superpositions of token embeddings obtained via averaging the embeddings of contiguous tokens in non-overlapping $s$-grams.
Each such superposed latent token, which we call \stokens, results in a single prediction, compared against the next non-overlapping bag-of-tokens.

The superposition objective is semi-causal and semi-autoregressive. In other words, the model still broadly predicts the input sequence from left to right, but we lose the ordering of the tokens in the superposition bag and the sampling capabilities during inference. Not surprisingly, if used as-is, a model trained using only TST has nonsensical outputs which represent a mixed probability of any of the $s$ future tokens. In order to remedy this, we opt for the simplest method which is to revert to training with the standard next-token causal prediction objective after some amounts of steps, where we define $r$ as the ratio of steps where we train with TST. In the second stage, we resume training from the saved checkpoint with the TST code fully removed to avoid any possibility of experimental and results contamination. 

We leave more complex conversion methods or other potential uses cases for TST as future work, where for example a model trained with TST might have informative latents that can be used to efficiently predict or verify multiple tokens at once, or using a TST-ed model as a compressive prior or encoder model. 

\subsection{Input Superposition: Bag-of-Token-Embeddings}

First, the contiguous tokenized data sequence of shape $B \times L \times V$ is segmented into \textbf{non-overlapping} contiguous segments of $s$ tokens, called bags. Here, $B$ denotes the batch size, $L$ the data token sequence length and $V$ the vocabulary size of the tokenizer. This gives us a bagged view of the data in the shape of $B \times l \times s \times V$, where $s$ is the superposition bag size and $l$ the latent \stoken sequence length.

In the embedding layer of the model, a single latent "s-token" is created by superposition of the tokens in a bag via the average of their embeddings, resulting in a final shape of $B \times l \times d$, where $d$ is the residual dimension of the model. The code is shown in Appendix~\ref{app:code}.

As the model performs computations over a coarser-grained representation of the input text, it processes $s$ times more data tokens for every FLOPs used on the latent \stokens. Therefore, in order to make a valid comparison during training, we choose to make every TST step equal-FLOPs to the baseline training by increasing the data sequence length $L$ by $s$ times during the superposition phase. 

This is what allows the model to ingest tokens at a higher rate during the superposition phase for the same amount of FLOPs per step compared to standard training. A faster alternative would be to increase the micro batch size by $s$ instead, but it is not explored in this work.

\subsection{Output Superposition: Next Bag-of-Tokens Prediction with Multi-hot Cross-Entropy Loss}
\label{sec:outputsuperposition}
Second, to predict the next "bag of tokens" instead of a single token with a single output head, we modify the standard one-hot cross-entropy (CE) loss into a \textbf{multi-hot} cross entropy (MCE) loss.

We have the standard CE loss with respect to the predicted logits $\mathbf{z}$ and the label index $y$:
\begin{equation}
\mathcal{L}_{\text{CE}}(\mathbf{z}, y) = -z_y + \log \sum_{i=1}^{V} \exp(z_i) 
\end{equation}

One possible expansion of this one-hot CE loss into a multi-hot MCE loss is simply to target an equal probability $1/s$ for each valid label (that together sum to 1). In our case, the label bag size $|\mathbf{y}|$ is equal to $s$. With some rearranging, we get the following:
\begin{equation}
    \mathcal{L}_{\text{MCE}}(\mathbf{z}, \mathbf{y}) = \frac{1}{|\mathbf{y}|}\sum_{y \in \mathbf{y}} \left(- z_y + \log \sum_{i=1}^{V} \exp(z_i)\right) - \log |\mathbf{y}|
\end{equation}

The expanded derivation is shown in Eq.~\ref{eq:lossderiv}, Appendix~\ref{app:derivloss}.

If we do not care about the absolute loss value during training and only care about the training dynamics, we can drop the $\log |\mathbf{y}|$ term from the loss, as it's gradient is 0. This yields the final simplified form that we use:
\begin{equation}
    \mathcal{L}_{\text{MCE}}(\mathbf{z}, \mathbf{y}) = \frac{1}{|\mathbf{y}|}\sum_{y \in \mathbf{y}} \mathcal{L}_{\text{CE}}(\mathbf{z}, y)
\end{equation}

Finally, the standard next token prediction labels are shifted to the left by $s-1$ before splitting into non-overlapping bags to preserve causality. This ensures that each bag of $s$ tokens at the positions $[t, t+s-1]$ predicts the next bag of tokens at $[t+s, t+2s-1]$.

The simplified form allows us to take advantage of the highly optimized and compiled CE loss kernels that exist in major pre-training libraries and use them without major modifications to the training code. Further optimization can be done to reuse the inner $\log \sum_i \exp(z_i)$ term, but that is not explored in this work as the overhead to summing outside the loss with a for loop is negligible and uses no additional memory. The code is shown in Appendix~\ref{app:code}.

We also tried many variants of possible multi-hot bag losses, such as Hinge loss~\citep{cortes1995support} and Binary Cross-Entropy (BCE) loss~\citep{shannon1948mathematical}. But these results were significantly worse than the MCE loss we picked and even worse than the baseline training without TST. Although an interesting alternative to the MCE loss shown here was explored, with more details in Appendix~\ref{app:derivloss-alt}, the final chosen MCE loss has the best balance between soundness and simplicity.

\section{Experiments}

\label{sec:exp}

We perform a battery of experiments that cover a wide range of superposition settings, varying the superposition bag size and ratio at different model scales and total durations. For all trainings, the TorchTitan~\citep{liang_torchtitan_2024} pre-training library was used with FSDP~\citep{zhao_pytorch_2023} parallelism, running on 64 NVIDIA B200 GPUs for the bigger models, and 8 B200 GPUs for the smaller models.

For pre-training the smaller models, we broadly follow the training procedures as outlined in SmolLM~\citep{allal_smollm2_2025}. For the 270M and 600M parameter models, we adapt and use the shape of the SmolLM2 models to fit the Llama3~\citep{grattafiori_llama_2024} modeling code, with two modifications: we use the Llama3-8B tokenizer, and we do not tie the weights of the input embeddings to the output head.
All other settings were kept the same (including layers, LM heads, inner dimensions, etc.).
The untied embeddings make our models bigger, with the 270M matching SmolLM2-135M and the 600M matching SmolLM2-360M.
Similarly for the 3B run, we matched it with SmolLM3-3B with the same modifications applied.
Finally, the DCLM~\citep{li2024datacomplm} dataset was used as the pre-training dataset, with a standard batch size of 2M tokens\footnote{or \stokens in the case of TST, in which we call both "equivalent-tokens" in the figures.} for the SmolLM-like models. 
More detail is shown in Table~\ref{tab:results}.

\begin{figure*}[ht]
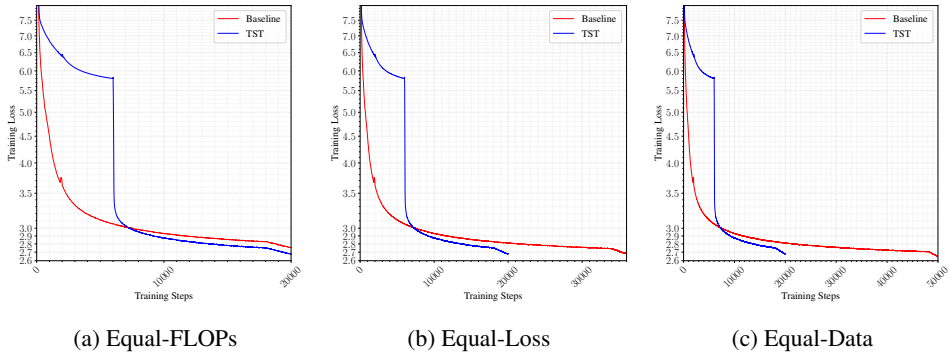

    \centering
    \begin{subfigure}[t]{0.3\linewidth}
        \centering
        \resizebox{\linewidth}{!}{\input{figures/isoflops.pgf}}
        \caption{Equal-FLOPs}
        \label{fig:isoflops}
    \end{subfigure}
    \begin{subfigure}[t]{0.3\linewidth}
        \centering
        \resizebox{\linewidth}{!}{\input{figures/isoloss.pgf}}
        \caption{Equal-Loss}
        \label{fig:isoloss}
    \end{subfigure}
    \begin{subfigure}[t]{0.3\linewidth}
        \centering
        \resizebox{\linewidth}{!}{\input{figures/isodata.pgf}}
        \caption{Equal-Data}
        \label{fig:isodata}
    \end{subfigure}
    \caption{Same constraints comparisons between different baseline training and one Token Superposition Training, using superposition bag size $s=6$ and step ratio $r=0.3$. These are the four 3B parameter runs described in Table~\ref{tab:results}.}
    \label{fig:iso-figures}
\end{figure*}

\begin{table*}[h]
\centering
\begin{adjustbox}{max width=\textwidth}
\begin{tabular}{lccccccc|ccccc}
\toprule
\textbf{Model} & \textbf{Params} & \textbf{TST} & \textbf{Total} & \textbf{TST} & \textbf{TST}   & \textbf{Total} & \textbf{B200-Hours} & \textbf{Final Loss} & \textbf{HellaSwag} & \textbf{ARC-E} & \textbf{ARC-C} & \textbf{MMLU} \\
\textbf{} & \textbf{} & \textbf{Steps} & \textbf{Steps} & \textbf{Bag Size} & \textbf{Tokens}  & \textbf{Tokens} & \textbf{($\downarrow$)} & \textbf{($\downarrow$)} & \textbf{($\uparrow$)} & \textbf{($\uparrow$)} & \textbf{($\uparrow$)} & \textbf{($\uparrow$)} \\
\midrule
Dense Baseline  & 270M  & --    & 20000  & -- & --   & 42B  & 34  & 3.212 & 36.3 & 46.7 & 24.9 & -- \\
Dense TST       & 270M  & 6000  & 20000  & 6x & 75B  & 105B & 34  & \textbf{3.142} & \textbf{38.6} & \textbf{47.6} & \textbf{26.4} & -- \\
\midrule
Dense Baseline  & 270M  & --    & 100000 & -- & --   & 209B & 170 & 3.092 & 40.2 & 47.5 & \textbf{26.2} & -- \\
Dense TST       & 270M  & 30000 & 100000 & 6x & 377B & 524B & 170 & \textbf{3.048} & \textbf{42.6} & \textbf{50.3} & 25.5 & -- \\
\midrule
Dense Baseline  & 600M  & --    & 20000  & -- & --   & 42B  & 61  & 3.019 & 43.5 & 51.7 & 25.5 & -- \\
Dense TST       & 600M  & 6000  & 20000  & 6x & 75B  & 105B & 61  & \textbf{2.943} & \textbf{48.2} & \textbf{52.5} & \textbf{26.9} & -- \\
\midrule
Dense Baseline  & 3B    & --    & 20000  & -- & --   & 42B  & \textbf{247} & 2.808 & 57.6 & 60.6 & 31.9 & 31.2 \\
Dense Baseline  & 3B    & --    & 36000  & -- & --   & 75B  & 443 & 2.677 & 62.3 & 65.9 & 34.9 & 32.7 \\
Dense Baseline  & 3B    & --    & 50000  & -- & --   & 105B & 622 & \textbf{2.640} & \textbf{63.9} & \textbf{67.3} & \textbf{36.8} & \textbf{33.3} \\
Dense TST       & 3B    & 6000  & 20000  & 6x & 75B  & 105B & \textbf{247} & \underline{2.676} & \underline{62.4} & \underline{66.3} & \underline{36.0} & \underline{32.8} \\
\midrule
MoE Baseline    & 10B A1B & --    & 125000 & -- & -- & 1.05T & 12311 & 2.252 & 70.1 & 73.8 & 46.3 & 37.4 \\
MoE TST         & 10B A1B & 12483 & 49983  & 16x & 1.68T & 2T & \textbf{4768} & \textbf{2.236} & \textbf{71.2} & \textbf{74.2} & \textbf{47.3} & \textbf{39.0} \\
\bottomrule
\end{tabular}
\end{adjustbox}
\caption{Overview of TST's advantage across different configurations compared to standard training (baseline). The TST Tokens denote raw data token count before compression. All evals are 0-shot. More results are shown in Table~\ref{tab:results-expanded}, Appendix~\ref{app:additionalresults}.}
\label{tab:results}
\end{table*}

\begin{figure}[h]
    \centering
    \includegraphics[width=1.0\linewidth]{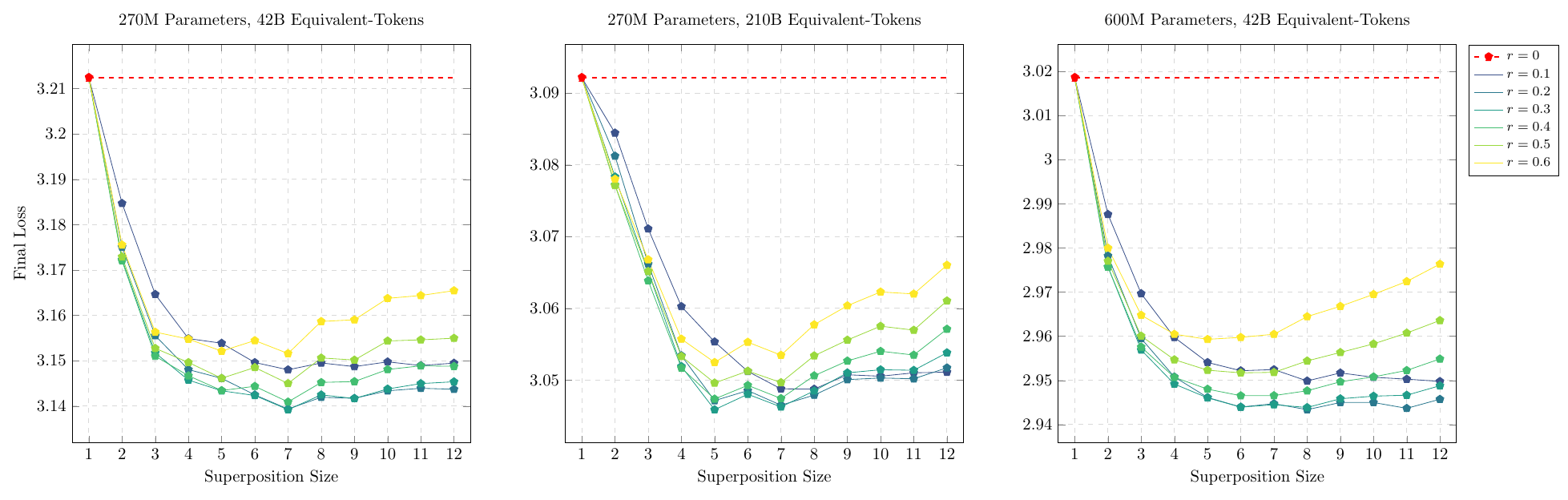}
    \caption{Superposition results with respect to loss at varying superposition bag sizes and superposition step ratio $r$, where $r$ is the ratio of number of steps trained in the superposition regime, with $1-r$ trained in the normal regime. Each data point is a fully trained model using TST first and converted to a standard AR LM. The full set of results are attached in Appendix~\ref{app:additionalresults}.}
    \label{fig:superposition-main-results}
\end{figure}

\begin{figure}[h]
    \centering
    \includegraphics[width=1.0\linewidth]{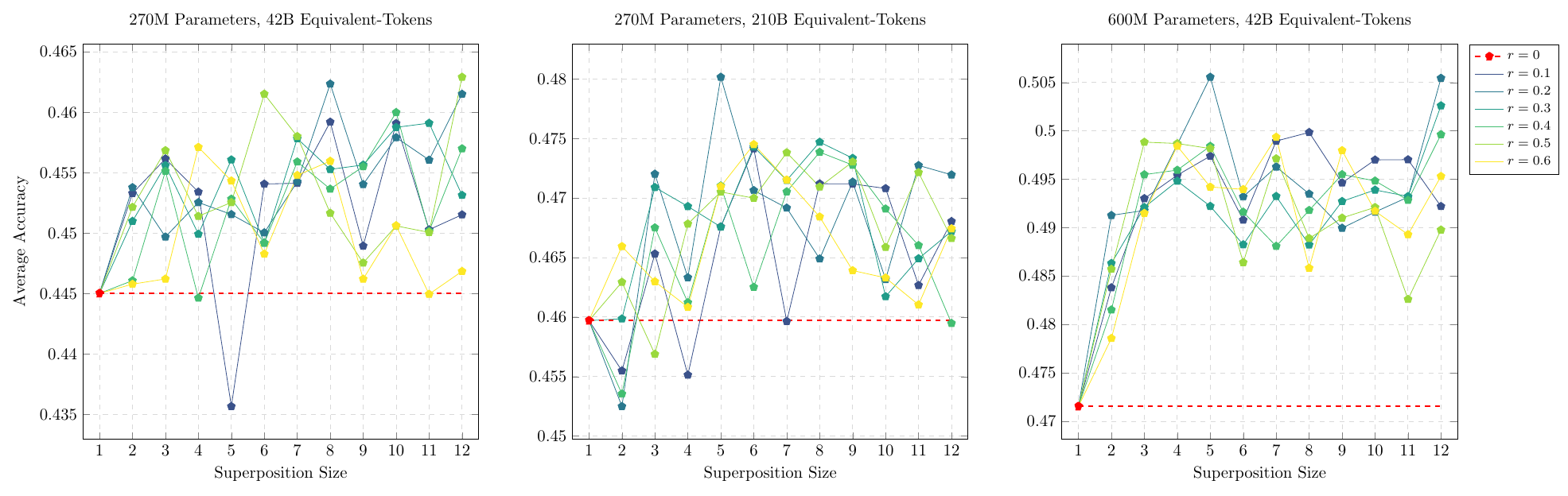}
    \caption{Downstream evals at varying superposition bag sizes and superposition step ratio $r$, where the average is the arithmetic average of (arc-c, arc-e, boolq, hellaswag, openbookqa, piqa, winogrande). The full set of results are attached in Appendix~\ref{app:additionalresults}.}
    \label{fig:superposition-evals-results}
\end{figure}

For the large 10B Mixture-of-Experts scaling validation run, we broadly follow the Qwen3 ~\citep{yang_qwen3_2025} training procedures, and use the Qwen3 architecture, but scaled down to match the size of a 10B total, 1B active parameter model trained to 1.05T tokens. A 50/50 mix of FineWeb-Edu~\citep{lozhkov2024fineweb-edu} and DCLM was used for this larger training run, along with a constant batch size of 8M tokens per step. The TST variant was trained with the settings shown in Table~\ref{tab:results}, for a total of 2T data tokens.

For the optimizer, we use AdamW~\citep{loshchilov2017decoupled}, with the optimal learning rate for the 270M and 600M models found using a sweep shown in Figure~\ref{fig:superposition-lr-results}, Appendix~\ref{app:lrsweep}. For the 3B and 10B models, we use the recommended learning rate of $2\times 10^{-4}$ and $3\times 10^{-4}$ respectively, as a full sweep is too expensive at this scale. For all training runs, we use the standard $\beta_1 = 0.9$ and $\beta_2 = 0.95$, along with the Warmup-Stable-Decay~\citep{hu2024minicpmunveilingpotentialsmall, wen2024understandingwarmupstabledecaylearningrates} learning rate scheduler. We warm up for a constant $2000$ steps and then decay for the last $10\%$ steps.

We run the standard set of LLM evaluations at the final step checkpoint after the recovery phase, including ARC~\citep{clark2018think}, BoolQ~\citep{clark2019boolq}, HellaSwag~\citep{zellers2019hellaswag}, MMLU~\citep{hendrycks2020measuring}, OpenBookQA~\citep{mihaylov2018can}, PIQA~\citep{bisk2020piqa} and Winogrande~\citep{sakaguchi2021winogrande}. All evaluations are performed using the Eleuther AI LM-Eval harness~\citep{eval-harness}, using a 0-shot prompting setting. These experiments are illustrated in Table~\ref{tab:results} and Figure~\ref{fig:superposition-main-results} highlights the robustness and significant benefit of using TST, as it outperforms the baseline in all equal-FLOPs or equal-loss settings tested.

\section{Discussion}

\subsection{On Comparisons to Auxiliary-loss Methods}

One may wonder about how TST compares empirically to multi-token prediction (MTP)~\citep{gloeckle_better_2024} and related auxiliary-loss methods~\citep{liu_deepseek-v3_2024, liu_next_2026, mahajan_beyond_2025, zuhri_predicting_2026}. We argue that such comparisons are not directly meaningful for our setting. MTP and its variants do not increase training-time throughput: they process the same number of tokens per FLOP as the baseline, while adding parameters and an auxiliary loss term.
Additionally, in the case of MTP and some of the related methods, emphasis is placed on inference-time gains via speculative decoding, an aspect that is not addressed in this work.
TST occupies a different point in the design space: it strictly increases tokens-per-FLOP during training, leaves the inference-time architecture untouched, and we observe consistent gains from 270M to 10B parameters.
We therefore view TST as orthogonal to, rather than competing with, auxiliary-loss methods, and combining the two is a natural direction for future work.

\subsection{Input and Output Superposition}
\label{sec:inputoutput}
We perform ablations adding the input-only, output-only and full superposition settings with a superposition bag size $s=4$ for a ratio $r=0.5$ of the total steps. The input-only method only bags the input token embeddings but only predicts one next token, and the output-only method processes individual tokens in the input but predicts the next bag of tokens in the output.

\begin{figure}[h]
    \centering
    \resizebox{0.5\linewidth}{!}{\input{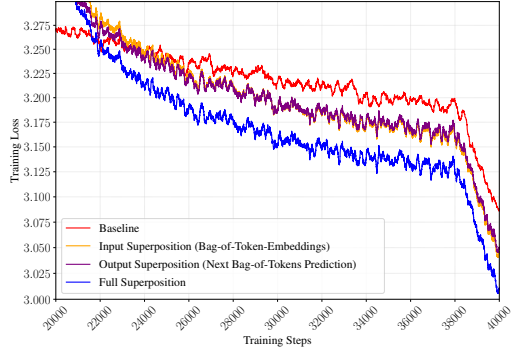}}
    \caption{Input and Output Superposition ablations, only the recovery phase (ii) is represented.}
    \label{fig:superposition-io-ablations}
\end{figure}

Figure \ref{fig:superposition-io-ablations} illustrates that all superposition settings outperform the baseline, but emphasizes that input or output superposition alone does not capture the complete improvement of full superposition, and the combination of both yields a further gain without signs of interference.
We understand this as evidence that TST is not a single trick but two orthogonal mechanisms: the input superposition changes the input granularity and FLOPs cost per unit of information, while the output side modifies the prediction target and the gradients.

\paragraph{Output Superposition}
Next bag-of-tokens prediction is using bag-of-words summaries of the future, reminiscent of seminal works in word vector representation~\citep{mikolov_efficient_2013}, extractive summarization~\citep{luhn_automatic_1958} and information retrieval~\citep{sparck_jones_statistical_1972}, repurposed as a local supervision signal for an autoregressive model. 
The closest work in the literature is \emph{future summary prediction}~\citep{mahajan_beyond_2025}, which also targets a compressed view of the
future.
The important difference is architectural rather than conceptual: the authors attach an auxiliary head with a binary cross-entropy loss on top of the regular next-token objective, paying extra parameters and an extra loss term.
We keep a single cross-entropy loss on the single main head and only replace the target.
An even closer match, to our knowledge not in the peer-reviewed literature, is a \texttt{modded-nanogpt} speedrun entry~\citep{kellerjordan_new_nodate} that concurrently proposed a next-bag-of-tokens loss.
It differs from ours in two design choices: exponential weighting of the bag, and a smooth interpolation into next-token prediction rather than a hard switch.

\paragraph{Output Bag Weighting}
Our experiments show U-shaped loss plots when varying the superposition bag size, highlighting the need to correctly tune this hyperparameter to find the optimal setting, with the risk of slightly suboptimal results if overshooting.
We tried different weighting schemes to average the loss participation of each term in the bag-of-tokens, detailed in Appendix \ref{app:weightloss}.
The power-law weighting scheme was the most promising at large superposition sizes, so we performed a new set of experiments and compared it to the uniform average and reported the results in Figure~\ref{fig:superposition-invrank-results}, Appendix~\ref{app:weightloss}.
The power-law weighted average results in higher loss than the uniform average for smaller superposition sizes, but outperforms it and is more stable  for $s\geq 8$.

\paragraph{Input Superposition}
Input superposition has, as far as we know, little direct analog in the LLM pretraining literature.
The underlying reason for its success is an open question.
One interpretation we find plausible is that the first phase acts as a form of pre-pre-training as studied by~\citet{hu_between_2025} and~\citet{lee_training_2026}: before learning full-resolution language, the model is exposed to a simpler distribution that already shares coarse statistical structure with natural language (\textit{e.g.} local topic, co-occurrence), and carries that inductive prior into phase~(ii).

A second, not exclusive, interpretation is that averaging in embedding space implicitly regularizes the embedding geometry, since many random $s$-grams must remain linearly separable once summed.
We do not have the interpretability evidence to pick between these.

Beyond language models, training first on a lower-resolution, higher-throughput version of the data and then on the full-resolution version is an old idea that recurs in recent work: \citet{anagnostidis_navigating_2024} schedule patch size from coarse to fine for Vision Transformers, \citet{minixhofer_bolmo_2025} start from a pretrained subword-LLM that is then converted into a model capable of processing finer-grained UTF-8 bytes.

Input superposition can be read as the same principle applied to token embeddings.
We see this as evidence that the coarse-to-fine granularity schedule is a reusable ingredient of efficient pretraining recipes, rather than a property of any specific modality.

\subsection{Two-Phase Task Alignment}
\label{sec:twophasealign}
The natural question arises: why has this effect not been observed in numerous prior work that resembles TST? Much of the existing literature on multi-stage pre-training and transfer learning relies on an \textbf{alignment phase}, in which the main model is frozen and only a small adapter is trained before unfreezing everything the recovery phase. We argue that this design choice is precisely what has obscured the effect we exploit in TST.

We hypothesize that the internal circuitry of a LLM is highly sensitive to its input and output representations. TST is, to our knowledge, one of the few compressive LLM pre-training method in which the input embedding and output LM head are shared without modification across the superposition and recovery phases, avoiding the representational mismatch that previous methods incur (and that adapter-based alignment is designed to patch over).

To test this hypothesis, we run a Dense TST 3B experiment matching the setup in Table~\ref{tab:results}, but we randomly re-initialize the input embedding and output LM head at the start of the recovery phase. As shown in Table~\ref{tab:results-align}, perturbing the input/output representations between the two phases completely eliminates the gains of TST, and makes it even worse than the baseline training where the TST steps are completely wasted and do not contribute to the final model. Although not definitive, this result supports our hypothesis that representation alignment across the two phases is a key reason why TST succeeds where prior compressive approaches have required explicit alignment training.

\begin{table*}[t]
\centering
\begin{adjustbox}{max width=\textwidth}
\begin{tabular}{lccccccc|c}
\toprule
\textbf{Model} & \textbf{Params} & \textbf{TST} & \textbf{Total} & \textbf{TST} & \textbf{TST}   & \textbf{Total} & \textbf{B200-Hours} & \textbf{Final Loss} \\
\textbf{} & \textbf{} & \textbf{Steps} & \textbf{Steps} & \textbf{Bag Size} & \textbf{Tokens}  & \textbf{Tokens} & \textbf{($\downarrow$)} & \textbf{($\downarrow$)} \\
\midrule
Dense Baseline  & 3B    & --    & 20000  & -- & --   & 42B  & 247 & \underline{2.808}\\
Dense TST       & 3B    & 6000  & 20000  & 6x & 75B  & 105B & 247 & \textbf{2.676}\\
Dense TST w/ Randomization       & 3B    & 6000  & 20000  & 6x & 75B  & 105B & 247 & 2.938\\
\bottomrule
\end{tabular}
\end{adjustbox}
\caption{Comparison between TST and TST with randomization where we re-initialize the input embedding and LM head layers between the superposition phase and the recovery phase.}
\label{tab:results-align}
\end{table*}

\section{Conclusion}

In this work, we describe a high throughput training paradigm for LLMs: Token Superposition Training.
During the token superposition regime, sample throughput is increased $s$-fold without changing per-step FLOPs, parallelism, model architecture, tokenizer, or data.
The following recovery phase is a return to the standard LLM pretraining regime, exhibiting a fast recovery period, quickly outperforming the loss of an equal-FLOPs baseline pretraining.
TST significantly increases the pretraining efficiency compared to baseline pretraining at the same computational cost (\textit{c.f.} Figure \ref{fig:isoflops}).
Alternatively, the same loss can be achieved at around half the computational cost (\textit{c.f.} Figure \ref{fig:isoloss}).
Overall, we find this new paradigm to be robust to hyperparameter choice within a reasonable range (superposition bag size $s\in [\![ 4, 8 ]\!]$ and step ratio $r\in \left[ 0.2, 0.4 \right]$).

\section{Limitations, Future Work and Broader Impacts}

\label{sec:limits}

TST effectively trades more data consumption for better loss at a given computational cost.
The underlying assumption is that LLM pretraining is performed under compute-bound constraints rather than data-bound constraints.
\citet{kim_pre-training_2025} recently argued that this assumption will be wrong in the future, given current trends.
In this alternative view, output-only superposition offers a significant advantage, as it outperforms the baseline pretraining regime without increasing data consumption.
We leave the study of these settings and the comparison with auxiliary loss methods, such as MTP, for future work.

Folding the initial sequence into a sequence of bags of $s$ tokens results in a longer effective context during TST compared to the baseline regime.
This could likely have positive effects on long context performance that we did not evaluate, as there would be less truncation or splitting of native long context data, leaving this to future work.

We also did not perform larger scale ablations or multiple identical runs to evaluate statistical significance due to limited compute resources.
Future work could investigate scaling laws of token superposition, in order to predict the best TST settings for larger model sizes, including industry-scale pretraining.

We proposed some hypotheses on the phenomena involved in TST, but further interpretability work on this subject could definitely improve the understanding of the underlying mechanisms and the ramifications of token superposition.

\paragraph{Broader Impacts:} Our work improves the efficiency of large language model pre-training, reducing computational cost and energy usage, which may improve accessibility to a broader range of researchers. However, increased efficiency may also accelerate the development and deployment of such models, potentially amplifying known risks including misuse for generating harmful content and concerns around bias, fairness, and privacy. While our contribution is methodological and does not directly introduce new capabilities, it may indirectly increase the scale at which these systems are trained and used. We encourage future work to pair efficiency improvements with advances in safety and responsible deployment.

\bibliography{superposition}

\appendix

\newpage
\section{Code}
\label{app:code}

\begin{lstlisting}[language=Python, caption=Input folding in Pytorch]

[...] # within train loop

if superposition_bag_size is not None and superposition_bag_size > 1:
    bs, seq = inputs.shape
    inputs = inputs.reshape(bs, seq//superposition_bag_size, superposition_bag_size)

\end{lstlisting}

\begin{lstlisting}[language=Python, caption=Bag-of-Token embeddings input in Pytorch]

[...] # within model forward

            # Using superposition
            if len(tokens.shape) == 3:
                bs, sp_seq, superposition_bag_size = tokens.shape
                
                # Sum in float32 for better numerical precision
                h = self.tok_embeddings(tokens[..., 0])
                h_dtype = h.dtype
                h = h.float()
                for i in range(1, superposition_bag_size):
                    h = h + self.tok_embeddings(tokens[..., i]).float()
                
                h = (h / superposition_bag_size).to(h_dtype)
                
            else:
                h = self.tok_embeddings(tokens)


\end{lstlisting}

\begin{lstlisting}[language=Python, caption=Next bag-of-words prediction loss code in Pytorch]
import torch

def cross_entropy_loss(pred: torch.Tensor, labels: torch.Tensor) -> torch.Tensor:
    
    # Compute shapes
    bs, seq, dim = pred.shape
    label_bs, label_seq = labels.shape
    superposition_bag_size = label_seq // seq
    superposition_offset = superposition_bag_size - 1
    
    # Pre-flatten and perform causal padding
    pred = pred.flatten(0, 1).float()
    labels = torch.nn.functional.pad(labels, (0, superposition_offset), mode='constant', value=-100)[..., superposition_offset:].view((bs, seq, superposition_bag_size))

    # Compute loss
    loss = 0.
    w_total = 0.
    for i in range(superposition_bag_size):
        w = 1 # uniform weighting
        target = labels[..., i].flatten(0, 1)
        loss += w * torch.nn.functional.cross_entropy(pred, target)
        w_total += w
        
    return loss / w_total
    
\end{lstlisting}

\section{Learning rate sweeps}
\label{app:lrsweep}
\begin{figure}[H]
    \centering
    \includegraphics[width=1.0\linewidth]{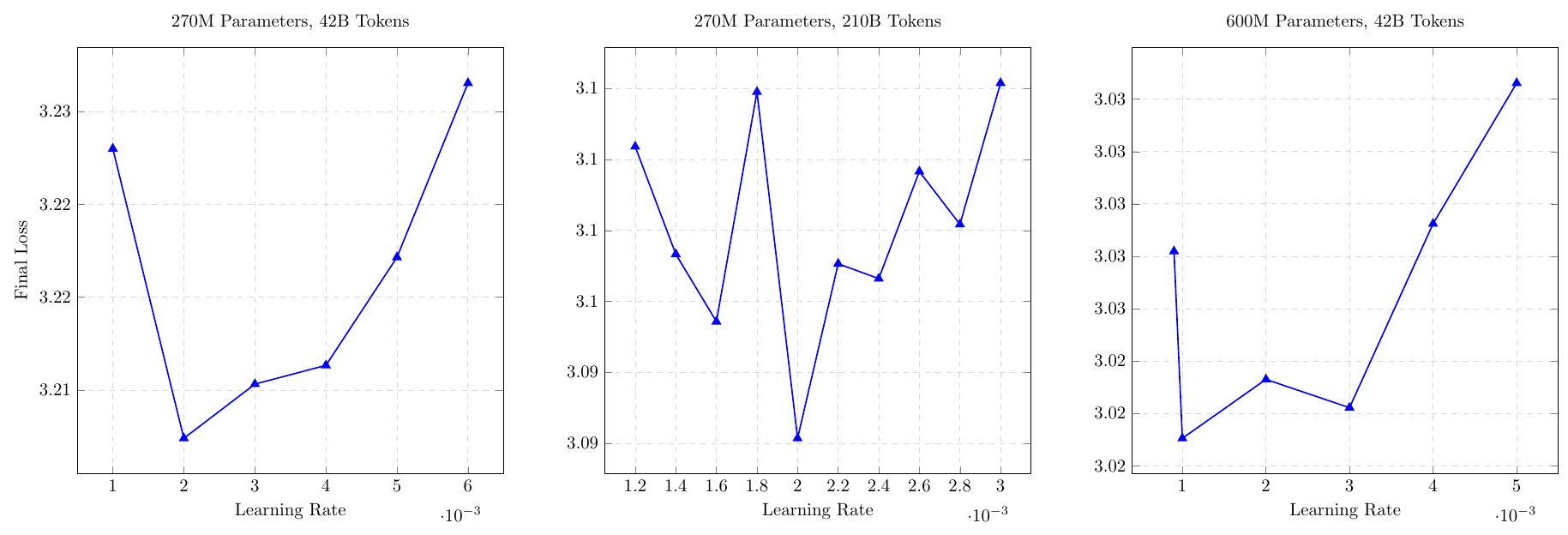}
    \caption{Learning rate sweeps at varying model sizes, with the optimal learning rate being used for all the training runs. The final lr used in order from left to right are $2\times 10^{-3}$, $2\times 10^{-3}$ and $1\times 10^{-3}$.}
    \label{fig:superposition-lr-results}
\end{figure}

\section{Loss Derivations}
\label{app:derivloss}

\paragraph{Setup.} Given logits $\mathbf{z} \in \mathbb{R}^V$, let $Z = \sum_{i=1}^V \exp(z_i)$ and write the softmax probability as $P(i) = \exp(z_i)/Z$. For a target distribution $\mathbf{t}$ over the vocabulary, cross-entropy and KL divergence are
$$\text{CE}(\mathbf{t}, P) = -\sum_i t_i \log P(i), \qquad \text{KL}(\mathbf{t}\,\|\,P) = \text{CE}(\mathbf{t}, P) - H(\mathbf{t}),$$
where $\mathbf{t}$ is the target probability vector, and where $H(\mathbf{t}) = -\sum_i t_i \log t_i$. Cross-entropy vanishes at its minimum only when $H(\mathbf{t}) = 0$, i.e. when $\mathbf{t}$ is one-hot. KL divergence always vanishes at $P = \mathbf{t}$ regardless of the target.

We now consider a bag $\mathbf{y}$ of $s$ valid target tokens, with two variants of a multi-hot cross-entropy (MCE) loss for a bag of $s$ valid target tokens $\mathbf{y}$.

\subsection{MCE: equal-probability targets}

The MCE loss used throughout the paper treats the bag as a multi-hot target and pushes each valid token's probability towards $1/s$, so that the bag tokens try to end up with an equal probability. Therefore, the target is the uniform distribution over the bag:
$$t_y = \begin{cases} 1/s & y \in \mathbf{y} \\ 0 & \text{otherwise}\end{cases}$$
Standard cross-entropy with this new target is
$$\text{CE}(\mathbf{t}, P) = -\frac{1}{s}\sum_{y\in\mathbf{y}} \log \frac{\exp(z_y)}{\sum_{i=1}^V \exp(z_i)} = -\frac{1}{s}\sum_{y\in\mathbf{y}} \log P(y)$$
Unlike the one-hot case, this target has nonzero entropy $H(\mathbf{t}) = \log s$, so plain CE bottoms out at $\log s$ rather than $0$. To recover the same "vanishes at the optimum" behaviour as standard CE, we subtract the entropy of the target (i.e. we use KL divergence):
\begin{equation}
\begin{aligned}
\mathcal{L}_{\text{MCE}}(\mathbf{z}, \mathbf{y})
&= \text{KL}(\mathbf{t}\,\|\,P) \\
&= -\frac{1}{s}\sum_{y\in\mathbf{y}} \log P(y) - \log s \\
&= -\frac{1}{s}\sum_{y\in\mathbf{y}} \log \frac{s\,\exp(z_y)}{\sum_{i=1}^V \exp(z_i)} \\
&= -\frac{1}{s} \sum_{y \in \mathbf{y}} \left(z_y - \log \sum_{i=1}^{V} \exp(z_i) + \log s \right) \\
&= -\frac{1}{s} \left(\sum_{y \in \mathbf{y}} z_y - s\log \sum_{i=1}^{V} \exp(z_i) + s \log s \right) \\
&=  - \frac{1}{s}\sum_{y\in\mathbf{y}} z_y + \log \sum_{i=1}^V \exp(z_i) - \log s
\end{aligned}
\label{eq:lossderiv}
\end{equation}

Rearranging the terms, we get:
\begin{equation}
\begin{aligned}
\mathcal{L}_{\text{MCE}}(\mathbf{z}, \mathbf{y})
&= -\frac{1}{|\mathbf{y}|}\sum_{y\in\mathbf{y}} z_y + \log \sum_{i=1}^{V} \exp(z_i) - \log |\mathbf{y}|\\
&= \frac{1}{|\mathbf{y}|} \sum_{y \in \mathbf{y}} \left(-z_y + \log \sum_{i=1}^{V} \exp(z_i) \right) - \log |\mathbf{y}| \\
&= \frac{1}{|\mathbf{y}|}\sum_{y \in \mathbf{y}} \mathcal{L}_{\text{CE}}(\mathbf{z}, y) - \log |\mathbf{y}|
\end{aligned}
\label{eq:lossrearrange}
\end{equation}

\subsection{MCE$_{\text{Alt}}$: sum-to-one probability targets}
\label{app:derivloss-alt}

We also investigated a different formulation of the MCE loss we used, where we target the sum of the probabilities of all valid labels to be 1 instead of targeting them to be equal-probability. Here we do not pick a single target distribution; instead we only require that the total probability mass on the bag be $1$. This effectively lets the model choose its own weighting across bag tokens. A natural way to express this is to treat the bag as a single composite label with probability
$$P(\mathbf{y}) = \sum_{y\in\mathbf{y}} P(y) = \frac{\sum_{y\in\mathbf{y}} \exp(z_y)}{\sum_{i=1}^V \exp(z_i)}.$$
Cross-entropy against a one-hot target on this composite label is
\begin{equation}
\begin{aligned}
\mathcal{L}_{\text{MCE}_{\text{Alt}}}(\mathbf{z}, \mathbf{y})
&= -\log P(\mathbf{y}) \\
&= -\log \frac{\sum_{y\in\mathbf{y}} \exp(z_y)}{\sum_{i=1}^V \exp(z_i)} \\
&=  - \log \sum_{y\in\mathbf{y}} \exp(z_y) + \log \sum_{i=1}^V \exp(z_i)
\end{aligned}
\end{equation}
Because the implicit target is a one-hot over composite labels (all mass in "the bag"), its entropy is zero and no correction is needed: the loss vanishes exactly when $\sum_{y\in\mathbf{y}} P(y) = 1$, just like ordinary CE on a single label.

In all the limited small-scale experiments we have done with MCE$_{\text{Alt}}$, it seemed to perform identically\footnote{We trained models using one variant of MCE or the other, and we obtained the same final loss after the recovery phase if everything else is kept identical.} compared to the equal-probability loss of MCE. We therefore did not explore it further, since unlike MCE it does not reduce to a simplified form that uses CE, it would require a custom loss function, reduce the training speed, increase memory usage, and add unnecessary complexity to the final method. We leave a more thorough exploration of this variant for future work.


\section{Non-uniform Multi-hot Cross-Entropy}

\label{app:weightloss}

\begin{figure}[h]
    \centering
    \includegraphics[width=0.7\linewidth]{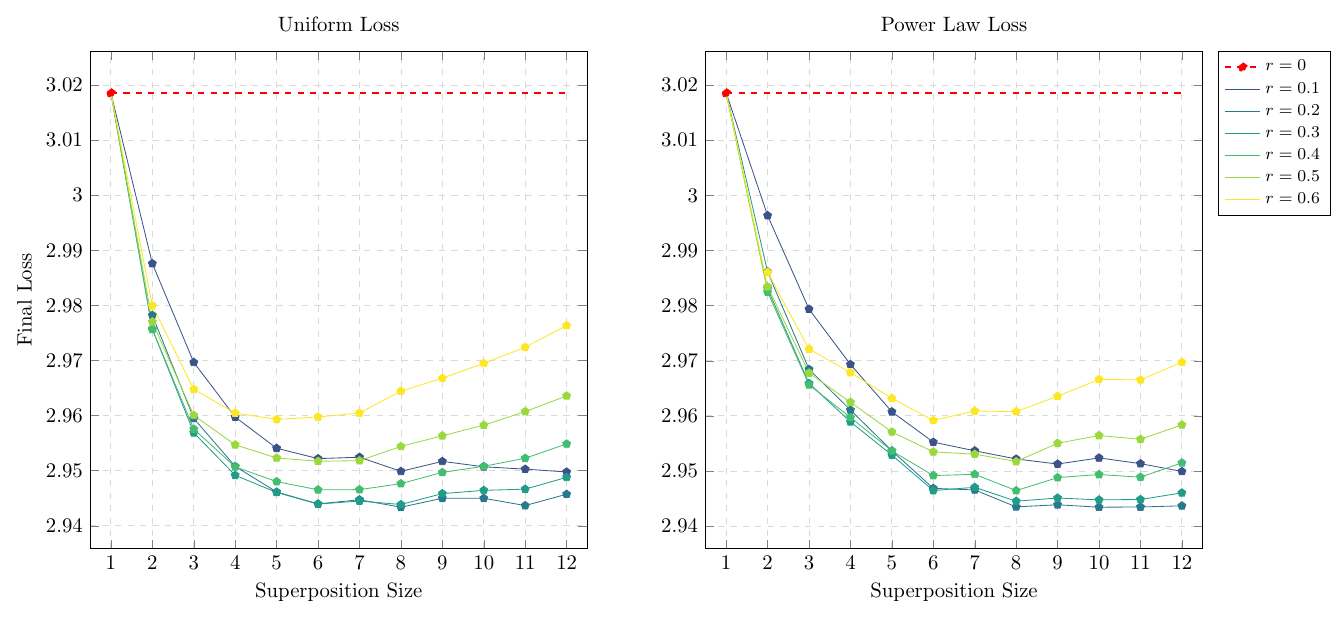}
    \caption{Comparison between superposition using uniform output loss and power law output loss at varying superposition window sizes and superposition ratio $r$, with a 600M-parameter model trained at 42B tokens. The final loss is the final training loss using standard next-token Cross-Entropy Loss.}
    \label{fig:superposition-invrank-results}
\end{figure}

For other multi-hot target distributions, we consider functions $g$ that are monotonically decreasing with the position $i$ in the bag.

\begin{equation}
\begin{aligned}
\mathcal{L}_{\text{MCE}}(\mathbf{z}, \mathbf{y}, g)
&= \frac{1}{\sum_{i}{g(i)}}\sum_{y \in \mathbf{y}} g(i)\mathcal{L}_{\text{CE}}(\mathbf{z}, y)
\end{aligned}
\label{eq:lossrearrange}
\end{equation}

\begin{itemize}
    \item Uniform: $i \mapsto 1$
    \item Power law: $i \mapsto \frac{1}{i}$
    \item Exponential: $i \mapsto \exp(-i)$, following \citep{kellerjordan_new_nodate}
    \item First token: $i \mapsto \delta_{1}({i})$
\end{itemize}

\begin{figure*}[ht]
    \centering
    \begin{subfigure}[t]{0.45\textwidth}
        \centering
        \includegraphics[width=\linewidth]{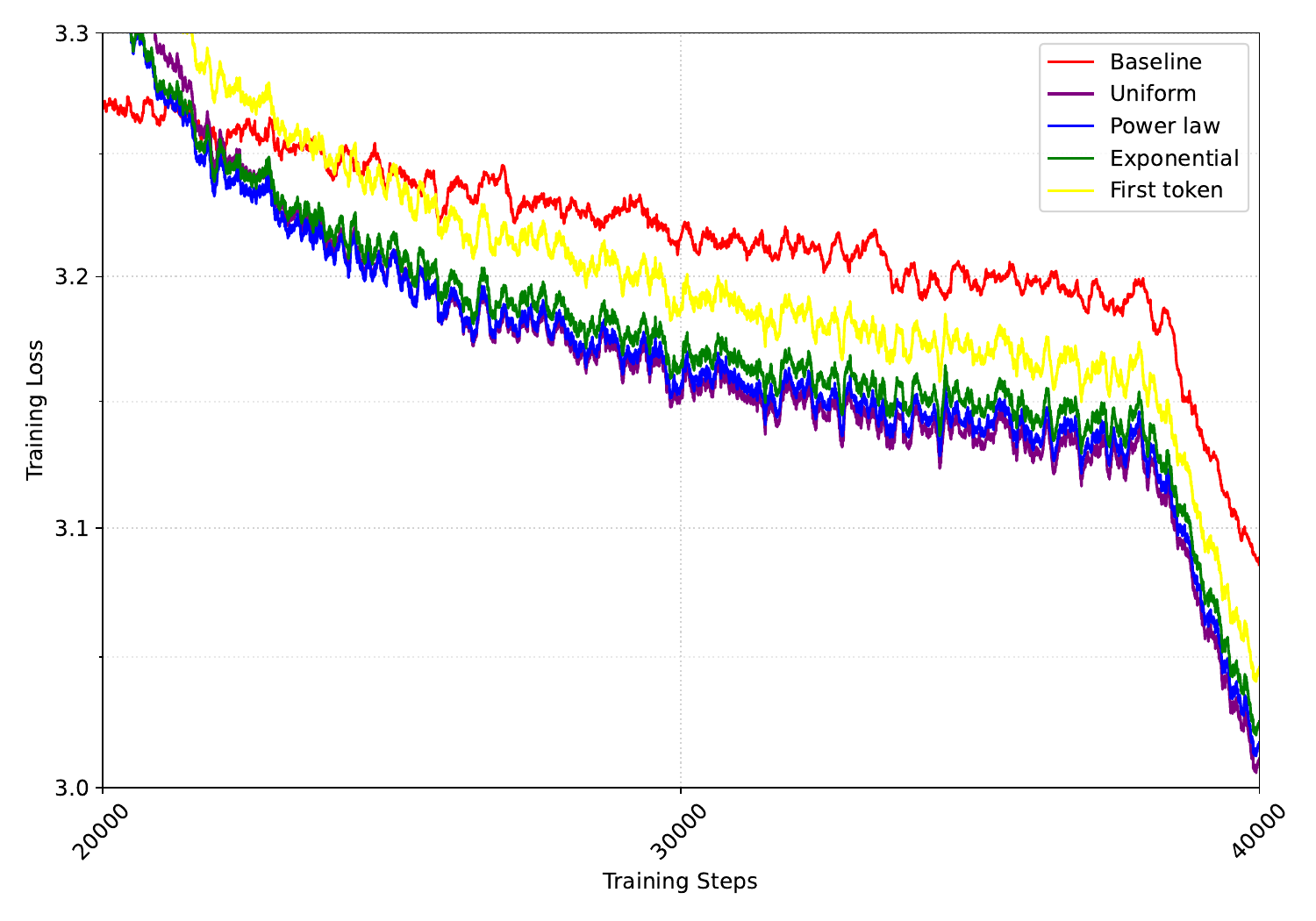}
        \caption{Superposition bag size $s=4$}
    \end{subfigure}
    \begin{subfigure}[t]{0.45\textwidth}
        \centering
        \includegraphics[width=\linewidth]{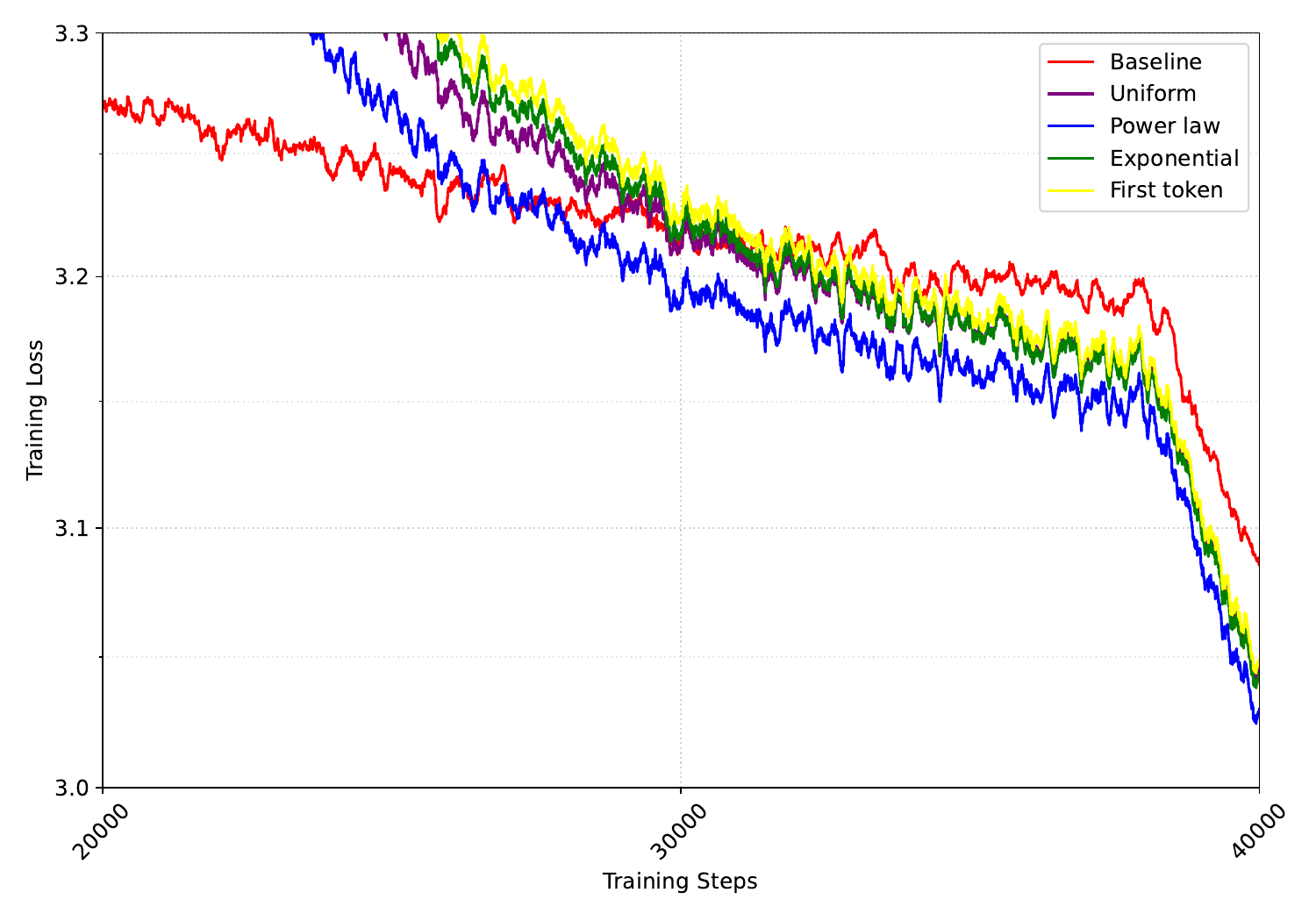}
        \caption{Superposition bag size $s=16$}
    \end{subfigure}
    \caption{Training loss after resuming from weighted superposition}
    \label{fig:loss-weighting-sup}
\end{figure*}

At a superposition bag size $s=16$, the best setting involves a power-law distribution.
However, at $s=4$, the uniform distribution works better.
We did not find a distribution that would work best for all superposition ratios.

\begin{figure}[H]
    \centering
    \includegraphics[width=0.8\linewidth]{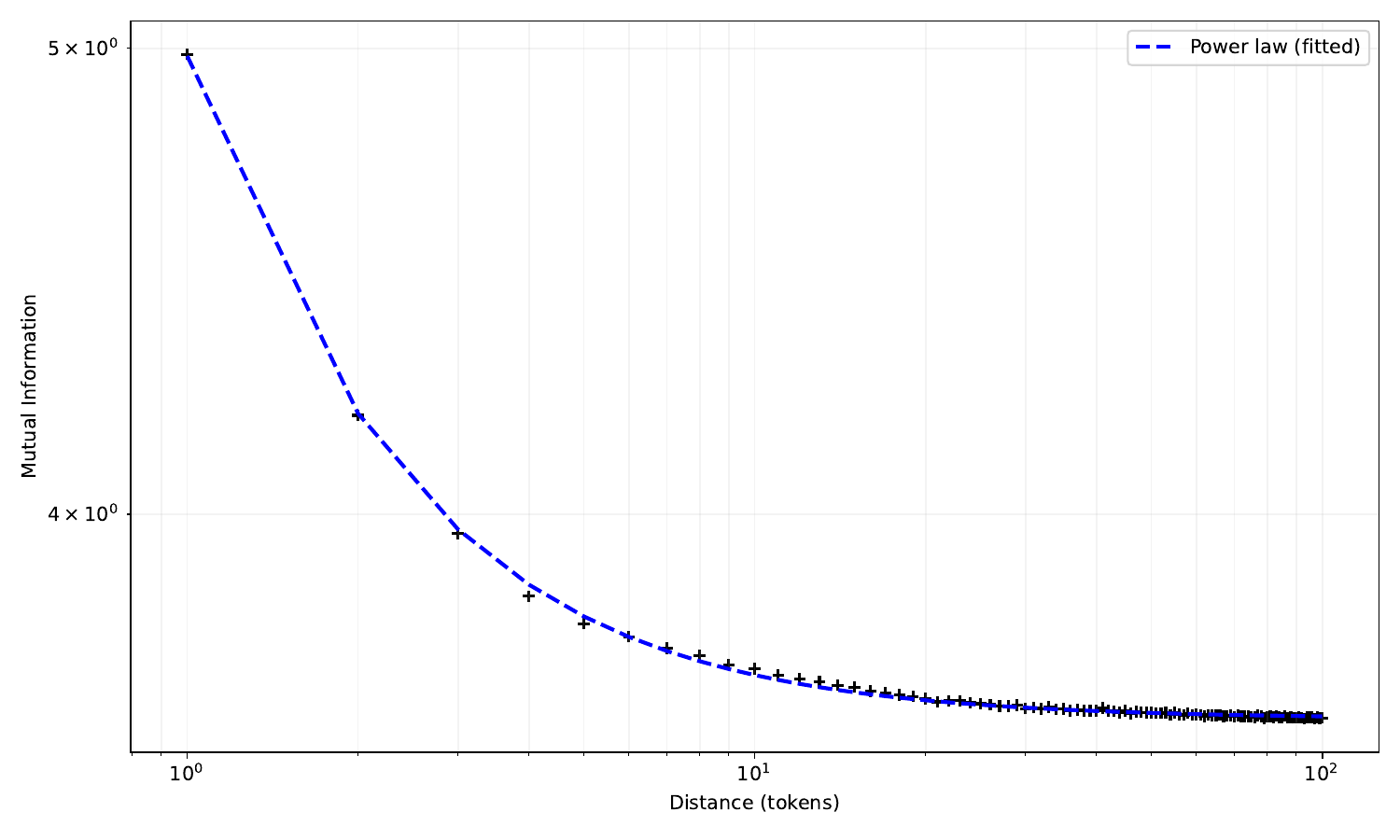}
    \caption{Mutual information between pairs of tokens in DCLM decays with distance following a power law: $d\mapsto C_0 + a * d^{k}$, with $C_0\approx3.63$, $a \approx 1.35$ and $k \approx -1.25$}
    \label{fig:loss-weighting}
\end{figure}

However, we believe that the power law distribution of losses is reminiscent of the prediction difficulty of future tokens based on current context.
\citet{ebeling_entropy_1994} investigated mutual information between pairs of letters in literary English texts, which they found to decay with distance following a power law.
Inspired by this, we compute mutual information between pairs of tokens sampled from texts in the DCLM dataset and fit a power law to modelize its decay.
This mutual information decay between pairs of tokens in the DCLM dataset, and a fitted power law, are illustrated in Figure~\ref{fig:loss-weighting}.
This result correlates with our observation for relative weighting of tokens for next bag-of-tokens prediction with large superposition bag sizes.
Using the values of the fitted power law to weight the losses per position results in a slightly better loss than the power law tested earlier.

\section{Additional Results}
Unless mentioned, all evals are run with 0-shot prompting.
\label{app:additionalresults}

\begin{figure}[h]
    \centering
    \begin{subfigure}{1.0\textwidth}
        \includegraphics[width=1.0\linewidth]{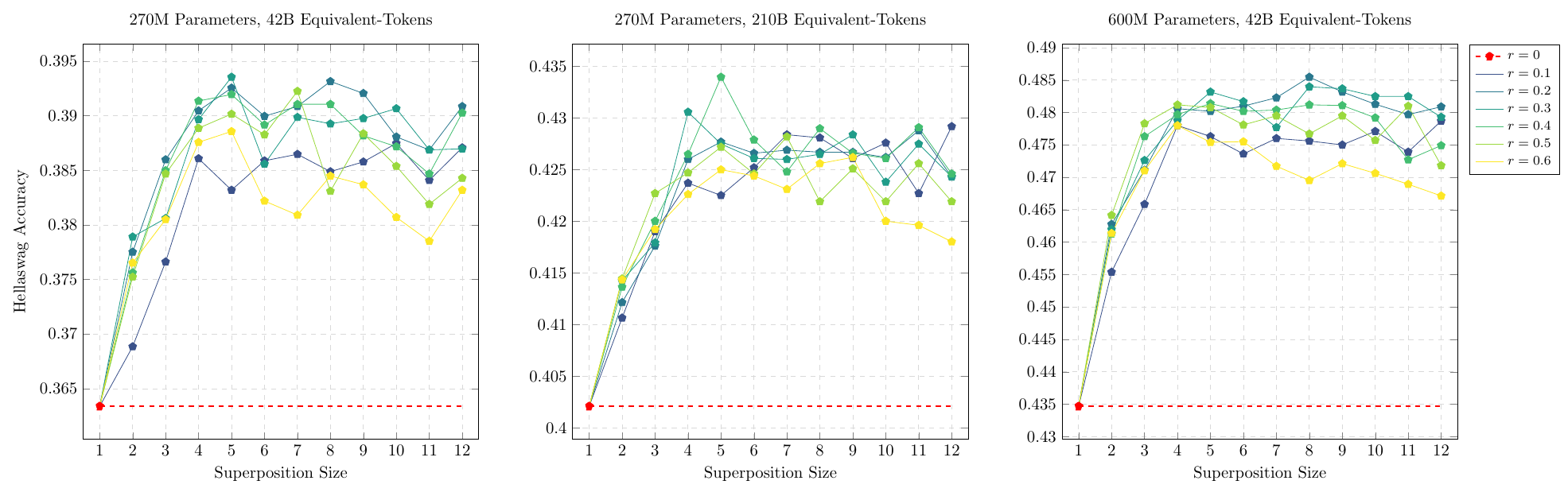}
    \end{subfigure}
    \hfill
    \begin{subfigure}{1.0\textwidth}
        \includegraphics[width=1.0\linewidth]{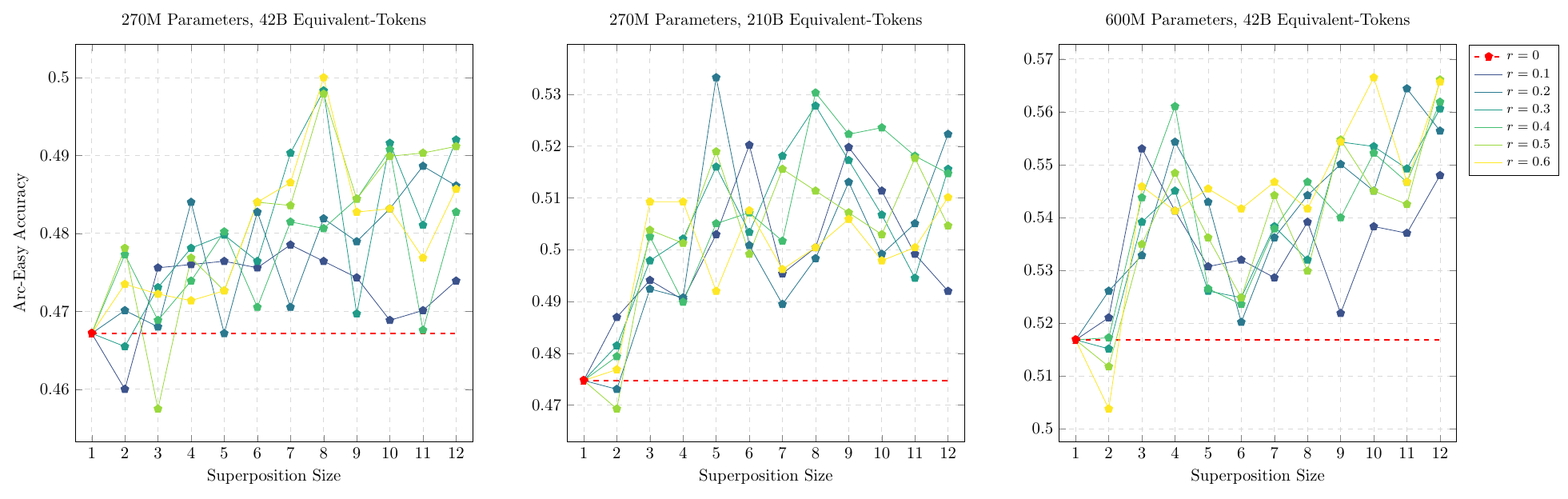}
    \end{subfigure}
    \hfill
    \caption{Rows from top to bottom: Hellaswag and ARC-Easy downstream evals at varying superposition bag sizes and superposition step ratio $r$.}
    \label{fig:superposition-evals-results-extra}
\end{figure}

\begin{table*}[t]
\centering
\begin{adjustbox}{max width=\textwidth}
\begin{tabular}{lcccc|ccccccccc}
\toprule
\textbf{Model} & \textbf{Params} & \textbf{TST} & \textbf{Total} & \textbf{B200-Hours} & \textbf{Final Loss} & \textbf{HellaSwag} & \textbf{ARC-E} & \textbf{ARC-C} & \textbf{MMLU} & \textbf{BoolQ} & \textbf{OpenBookQA} & \textbf{PIQA} & \textbf{Winogrande} \\
\textbf{} & \textbf{} & \textbf{Steps} & \textbf{Steps} & \textbf{($\downarrow$)} & \textbf{($\downarrow$)} & \textbf{($\uparrow$)} & \textbf{($\uparrow$)} & \textbf{($\uparrow$)} & \textbf{($\uparrow$)} & \textbf{($\uparrow$)} & \textbf{($\uparrow$)} & \textbf{($\uparrow$)} & \textbf{($\uparrow$)} \\
\midrule
Dense Baseline  & 270M  & --    & 20000  & 34  & 3.212 & 36.3 & 46.7 & 24.9 & -- & \textbf{56.7} & 29.2 & 66.4 & \textbf{51.3} \\
Dense TST       & 270M  & 6000  & 20000  & 34  & \textbf{3.142} & \textbf{38.6} & \textbf{47.6} & \textbf{26.4} & -- & 54.0 & \textbf{29.8} & \textbf{67.0} & 51.1 \\
\midrule
Dense Baseline  & 270M  & --    & 100000 & 170 & 3.092 & 40.2 & 47.5 & \textbf{26.2} & -- & \textbf{58.5} & 30.6 & 67.3 & 51.5 \\
Dense TST       & 270M  & 30000 & 100000 & 170 & \textbf{3.048} & \textbf{42.6} & \textbf{50.3} & 25.5 & -- & 57.9 & \textbf{32.4} & \textbf{69.0} & \textbf{54.2}\\
\midrule
Dense Baseline  & 600M  & --    & 20000  & 61  & 3.019 & 43.5 & 51.7 & 25.5 & -- & \textbf{56.6} & 31.2 & 69.0 & 52.6 \\
Dense TST       & 600M  & 6000  & 20000  & 61  & \textbf{2.943} & \textbf{48.2} & \textbf{52.5} & \textbf{26.9} & -- & 54.8 & \textbf{35.0} & \textbf{70.6} & \textbf{53.9} \\
\midrule
Dense Baseline  & 3B    & --    & 20000  & \textbf{247} & 2.808 & 57.6 & 60.6 & 31.9 & 31.2 & 58.4 & 36.6 & 74.5 & 56.5 \\
Dense Baseline  & 3B    & --    & 36000  & 443 & 2.677 & 62.3 & 65.9 & 34.9 & 32.7 & \underline{62.1} & 36.4 & 74.7 & \underline{60.0} \\
Dense Baseline  & 3B    & --    & 50000  & 622 & \textbf{2.640} & \textbf{63.9} & \textbf{67.3} & \textbf{36.8} & \textbf{33.3} & \textbf{64.6} & \underline{38.0} & \underline{75.5} & \textbf{60.9} \\
Dense TST       & 3B    & 6000  & 20000  & \textbf{247} & \underline{2.676} & \underline{62.4} & \underline{66.3} & \underline{36.0} & \underline{32.8} & 60.0 & \textbf{42.0} & \textbf{76.1} & 59.6 \\
\midrule
MoE Baseline    & 10B A1B & --    & 125000 & 12311 & 2.252 & 70.1 & 73.8 & 46.3 & 37.4 & 66.2 & \textbf{44.0} & 77.2 & 61.3 \\
MoE TST         & 10B A1B & 12483 & 49983  & \textbf{4768} & \textbf{2.236} & \textbf{71.2} & \textbf{74.2} & \textbf{47.3} & \textbf{39.0} & \textbf{69.4} & 43.2 & \textbf{77.4} & \textbf{63.0} \\
\bottomrule
\end{tabular}
\end{adjustbox}
\caption{Expanded results of Table~\ref{tab:results}. All evals are 0-shot.}
\label{tab:results-expanded}
\end{table*}

\begin{table}[htbp]
\centering
\caption{Final loss values with varying $r$ and $s$, for 270M models trained for 20k total steps. See Table~\ref{tab:results} for training details.}
\label{tab:data}
\resizebox{\textwidth}{!}{%
\begin{tabular}{c|cccccccccccccccc}
\hline
\diagbox{$r$}{$s$} & 1 & 2 & 3 & 4 & 5 & 6 & 7 & 8 & 9 & 10 & 11 & 12 & 13 & 14 & 15 & 16 \\
\hline
0.0 & 3.2124 & -- & -- & -- & -- & -- & -- & -- & -- & -- & -- & -- & -- & -- & -- & -- \\
0.1 & -- & 3.1847 & 3.1646 & 3.1549 & 3.1539 & 3.1496 & 3.1480 & 3.1495 & 3.1487 & 3.1498 & 3.1489 & 3.1495 & 3.1505 & 3.1513 & 3.1493 & 3.1487 \\
0.2 & -- & 3.1751 & 3.1555 & 3.1480 & 3.1461 & 3.1425 & 3.1393 & 3.1419 & 3.1417 & 3.1434 & 3.1439 & 3.1437 & 3.1450 & 3.1451 & 3.1431 & 3.1455 \\
0.3 & -- & 3.1723 & 3.1517 & 3.1457 & 3.1434 & 3.1423 & 3.1392 & 3.1425 & 3.1417 & 3.1438 & 3.1450 & 3.1454 & 3.1464 & 3.1475 & 3.1456 & 3.1484 \\
0.4 & -- & 3.1720 & 3.1510 & 3.1467 & 3.1434 & 3.1444 & 3.1409 & 3.1452 & 3.1454 & 3.1481 & 3.1489 & 3.1488 & 3.1504 & 3.1520 & 3.1508 & 3.1534 \\
0.5 & -- & 3.1729 & 3.1527 & 3.1496 & 3.1461 & 3.1485 & 3.1450 & 3.1506 & 3.1502 & 3.1544 & 3.1546 & 3.1550 & 3.1574 & 3.1597 & 3.1583 & 3.1611 \\
0.6 & -- & 3.1755 & 3.1563 & 3.1548 & 3.1521 & 3.1545 & 3.1516 & 3.1587 & 3.1590 & 3.1638 & 3.1644 & 3.1655 & 3.1675 & 3.1708 & 3.1699 & 3.1734 \\
0.7 & -- & 3.1807 & 3.1635 & 3.1630 & 3.1621 & 3.1664 & 3.1635 & 3.1720 & 3.1730 & 3.1794 & 3.1806 & 3.1823 & 3.1847 & 3.1893 & 3.1882 & 3.1922 \\
0.8 & -- & 3.1904 & 3.1768 & 3.1774 & 3.1808 & 3.1879 & 3.1862 & 3.1970 & 3.1998 & 3.2075 & 3.2105 & 3.2138 & 3.2176 & 3.2239 & 3.2226 & 3.2280 \\
0.9 & -- & 3.2131 & 3.2110 & 3.2221 & 3.2324 & 3.2487 & 3.2501 & 3.2681 & 3.2765 & 3.2894 & 3.2985 & 3.3073 & 3.3146 & 3.3275 & 3.3254 & 3.3381 \\
1.0 & -- & 4.6099 & 5.2482 & 5.6289 & 5.8658 & 6.0365 & 6.1567 & 6.2477 & 6.3146 & 6.3677 & 6.4152 & 6.4529 & 6.4835 & 6.5082 & 6.5314 & 6.5519 \\
\hline
\end{tabular}%
}
\end{table}
\begin{table}[htbp]
\centering
\caption{Final loss values with varying $r$ and $s$, for 270M models trained for 100k total steps. See Table~\ref{tab:results} for training details.}
\label{tab:data2}
\resizebox{\textwidth}{!}{%
\begin{tabular}{c|ccccccccccccc}
\hline
\diagbox{$r$}{$s$} & 1 & 2 & 3 & 4 & 5 & 6 & 7 & 8 & 9 & 10 & 11 & 12 & 13 \\
\hline
0.0 & 3.0921 & -- & -- & -- & -- & -- & -- & -- & -- & -- & -- & -- & -- \\
0.1 & -- & 3.0844 & 3.0711 & 3.0603 & 3.0553 & 3.0512 & 3.0488 & 3.0488 & 3.0508 & 3.0506 & 3.0510 & 3.0511 & 3.0525 \\
0.2 & -- & 3.0812 & 3.0662 & 3.0534 & 3.0472 & 3.0486 & 3.0465 & 3.0479 & 3.0501 & 3.0503 & 3.0502 & 3.0517 & 3.0524 \\
0.3 & -- & 3.0783 & 3.0652 & 3.0519 & 3.0459 & 3.0480 & 3.0463 & 3.0485 & 3.0510 & 3.0515 & 3.0514 & 3.0538 & 3.0551 \\
0.4 & -- & 3.0772 & 3.0639 & 3.0517 & 3.0474 & 3.0493 & 3.0475 & 3.0506 & 3.0527 & 3.0540 & 3.0535 & 3.0571 & 3.0588 \\
0.5 & -- & 3.0771 & 3.0652 & 3.0533 & 3.0496 & 3.0513 & 3.0497 & 3.0534 & 3.0556 & 3.0575 & 3.0570 & 3.0611 & 3.0627 \\
0.6 & -- & 3.0780 & 3.0668 & 3.0557 & 3.0525 & 3.0553 & 3.0535 & 3.0577 & 3.0604 & 3.0623 & 3.0620 & 3.0660 & 3.0681 \\
0.7 & -- & 3.0805 & 3.0703 & 3.0601 & 3.0574 & 3.0608 & 3.0597 & 3.0642 & 3.0673 & 3.0697 & 3.0694 & 3.0738 & 3.0762 \\
0.8 & -- & 3.0852 & 3.0770 & 3.0683 & 3.0670 & 3.0711 & 3.0706 & 3.0761 & 3.0800 & 3.0827 & 3.0830 & 3.0881 & 3.0909 \\
0.9 & -- & 3.0984 & 3.0963 & 3.0908 & 3.0934 & 3.1009 & 3.1020 & 3.1097 & 3.1161 & 3.1209 & 3.1236 & 3.1306 & 3.1348 \\
1.0 & -- & 4.5120 & 5.2018 & 5.5772 & 5.8255 & 5.9952 & 6.1179 & 6.2156 & 6.2891 & 6.3461 & 6.3890 & 6.4285 & 6.4628 \\
\hline
\end{tabular}%
}
\end{table}

\begin{table}[htbp]
\centering
\caption{Final loss values with varying $r$ and $s$, for 600M models trained for 20k total steps. See Table~\ref{tab:results} for training details.}
\label{tab:data}
\resizebox{\textwidth}{!}{%
\begin{tabular}{c|ccccccccccccc}
\hline
\diagbox{$r$}{$s$} & 1 & 2 & 3 & 4 & 5 & 6 & 7 & 8 & 9 & 10 & 11 & 12 & 13 \\
\hline
0.0 & 3.0186 & -- & -- & -- & -- & -- & -- & -- & -- & -- & -- & -- & -- \\
0.1 & -- & 2.9876 & 2.9697 & 2.9597 & 2.9541 & 2.9522 & 2.9525 & 2.9499 & 2.9517 & 2.9507 & 2.9503 & 2.9498 & 2.9514 \\
0.2 & -- & 2.9782 & 2.9595 & 2.9508 & 2.9461 & 2.9440 & 2.9447 & 2.9434 & 2.9450 & 2.9450 & 2.9437 & 2.9457 & 2.9465 \\
0.3 & -- & 2.9757 & 2.9569 & 2.9492 & 2.9461 & 2.9439 & 2.9445 & 2.9439 & 2.9458 & 2.9464 & 2.9467 & 2.9488 & 2.9501 \\
0.4 & -- & 2.9757 & 2.9576 & 2.9507 & 2.9480 & 2.9465 & 2.9466 & 2.9477 & 2.9497 & 2.9508 & 2.9523 & 2.9549 & 2.9558 \\
0.5 & -- & 2.9770 & 2.9601 & 2.9547 & 2.9523 & 2.9517 & 2.9518 & 2.9544 & 2.9563 & 2.9583 & 2.9607 & 2.9636 & 2.9647 \\
0.6 & -- & 2.9800 & 2.9648 & 2.9605 & 2.9593 & 2.9598 & 2.9605 & 2.9645 & 2.9668 & 2.9695 & 2.9724 & 2.9764 & 2.9779 \\
0.7 & -- & 2.9857 & 2.9723 & 2.9702 & 2.9708 & 2.9726 & 2.9747 & 2.9801 & 2.9832 & 2.9871 & 2.9911 & 2.9962 & 2.9987 \\
0.8 & -- & 2.9946 & 2.9865 & 2.9874 & 2.9918 & 2.9955 & 3.0000 & 3.0079 & 3.0132 & 3.0181 & 3.0243 & 3.0310 & 3.0356 \\
0.9 & -- & 3.0150 & 3.0194 & 3.0305 & 3.0441 & 3.0546 & 3.0659 & 3.0811 & 3.0915 & 3.1029 & 3.1146 & 3.1264 & 3.1366 \\
1.0 & -- & 4.3622 & 5.0131 & 5.4091 & -- & 5.8477 & 5.9880 & 6.0884 & 6.1684 & 6.2315 & 6.2853 & 6.3318 & 6.3679 \\
\hline
\end{tabular}%
}
\end{table}

\begin{table}[htbp]
\centering
\caption{Final loss values with varying $r$ and $s$, for 600M models trained for 20k total steps. See Table~\ref{tab:results} for training details. The TST loss used the power-law weighting as described in Appendix~\ref{app:weightloss}.}
\label{tab:data}
\resizebox{\textwidth}{!}{%
\begin{tabular}{c|ccccccccccccc}
\hline
\diagbox{$r$}{$s$} & 1 & 2 & 3 & 4 & 5 & 6 & 7 & 8 & 9 & 10 & 11 & 12 & 13 \\
\hline
0.0 & 3.0186 & -- & -- & -- & -- & -- & -- & -- & -- & -- & -- & -- & -- \\
0.1 & -- & 2.9963 & 2.9794 & 2.9693 & 2.9608 & 2.9552 & 2.9537 & 2.9522 & 2.9513 & 2.9524 & 2.9513 & 2.9499 & 2.9497 \\
0.2 & -- & 2.9861 & 2.9685 & 2.9611 & 2.9537 & 2.9469 & 2.9466 & 2.9435 & 2.9439 & 2.9434 & 2.9435 & 2.9437 & 2.9442 \\
0.3 & -- & 2.9831 & 2.9659 & 2.9589 & 2.9529 & 2.9465 & 2.9471 & 2.9445 & 2.9451 & 2.9448 & 2.9448 & 2.9460 & 2.9471 \\
0.4 & -- & 2.9825 & 2.9656 & 2.9597 & 2.9537 & 2.9492 & 2.9494 & 2.9465 & 2.9488 & 2.9494 & 2.9489 & 2.9515 & 2.9528 \\
0.5 & -- & 2.9834 & 2.9678 & 2.9625 & 2.9571 & 2.9535 & 2.9531 & 2.9517 & 2.9550 & 2.9564 & 2.9558 & 2.9584 & 2.9606 \\
0.6 & -- & 2.9860 & 2.9721 & 2.9679 & 2.9632 & 2.9592 & 2.9609 & 2.9608 & 2.9636 & 2.9666 & 2.9665 & 2.9697 & 2.9721 \\
0.7 & -- & 2.9908 & 2.9786 & 2.9766 & 2.9733 & 2.9707 & 2.9741 & 2.9752 & 2.9781 & 2.9824 & 2.9828 & 2.9865 & 2.9895 \\
0.8 & -- & 2.9988 & 2.9904 & 2.9922 & 2.9911 & 2.9909 & 2.9967 & 2.9988 & 3.0038 & 3.0098 & 3.0111 & 3.0163 & 3.0202 \\
0.9 & -- & 3.0162 & 3.0184 & 3.0277 & 3.0334 & 3.0403 & 3.0510 & 3.0579 & 3.0672 & 3.0785 & 3.0829 & 3.0925 & 3.0997 \\
1.0 & -- & 3.0669 & -- & -- & -- & -- & -- & -- & -- & -- & -- & -- & -- \\
\hline
\end{tabular}%
}
\end{table}

\begin{table}[htbp]
\centering
\caption{Final ARC-Challenge evals with varying $r$ and $s$, for 270M models trained for 20k total steps. See Table~\ref{tab:results} for training details.}
\label{tab:data}
\resizebox{\textwidth}{!}{%
\begin{tabular}{c|cccccccccccccccc}
\hline
\diagbox{$r$}{$s$} & 1 & 2 & 3 & 4 & 5 & 6 & 7 & 8 & 9 & 10 & 11 & 12 & 13 & 14 & 15 & 16 \\
\hline
0.0 & 0.2491 & -- & -- & -- & -- & -- & -- & -- & -- & -- & -- & -- & -- & -- & -- & -- \\
0.1 & -- & 0.2543 & 0.2509 & 0.2432 & 0.2491 & 0.2645 & 0.2517 & 0.2594 & 0.2432 & 0.2594 & 0.2526 & 0.2526 & 0.2526 & 0.2568 & 0.2440 & 0.2611 \\
0.2 & -- & 0.2500 & 0.2483 & 0.2526 & 0.2509 & 0.2457 & 0.2483 & 0.2517 & 0.2679 & 0.2560 & 0.2517 & 0.2628 & 0.2415 & 0.2509 & 0.2500 & 0.2602 \\
0.3 & -- & 0.2449 & 0.2602 & 0.2577 & 0.2534 & 0.2637 & 0.2509 & 0.2534 & 0.2568 & 0.2526 & 0.2543 & 0.2645 & 0.2577 & 0.2577 & 0.2491 & 0.2577 \\
0.4 & -- & 0.2466 & 0.2474 & 0.2398 & 0.2594 & 0.2637 & 0.2491 & 0.2483 & 0.2679 & 0.2509 & 0.2534 & 0.2739 & 0.2594 & 0.2628 & 0.2483 & 0.2551 \\
0.5 & -- & 0.2526 & 0.2713 & 0.2551 & 0.2491 & 0.2560 & 0.2594 & 0.2534 & 0.2500 & 0.2483 & 0.2517 & 0.2765 & 0.2449 & 0.2491 & 0.2483 & -- \\
0.6 & -- & 0.2449 & 0.2543 & 0.2594 & 0.2543 & 0.2551 & 0.2526 & 0.2577 & 0.2398 & 0.2619 & 0.2440 & 0.2637 & 0.2517 & 0.2568 & 0.2551 & 0.2449 \\
0.7 & -- & 0.2440 & 0.2474 & 0.2551 & 0.2440 & 0.2534 & 0.2526 & 0.2457 & 0.2389 & 0.2568 & 0.2551 & 0.2602 & 0.2483 & 0.2645 & 0.2602 & 0.2517 \\
0.8 & -- & 0.2423 & 0.2398 & 0.2585 & 0.2355 & 0.2483 & 0.2577 & 0.2594 & 0.2304 & 0.2662 & 0.2577 & 0.2594 & 0.2474 & 0.2543 & 0.2568 & 0.2483 \\
0.9 & -- & 0.2346 & 0.2440 & 0.2398 & 0.2517 & 0.2483 & 0.2585 & 0.2491 & 0.2440 & 0.2432 & 0.2389 & 0.2440 & 0.2526 & 0.2432 & 0.2398 & 0.2363 \\
\hline
\end{tabular}%
}
\end{table}

\begin{table}[htbp]
\centering
\caption{Final ARC-Easy evals with varying $r$ and $s$, for 270M models trained for 20k total steps. See Table~\ref{tab:results} for training details.}
\label{tab:data}
\resizebox{\textwidth}{!}{%
\begin{tabular}{c|cccccccccccccccc}
\hline
\diagbox{$r$}{$s$} & 1 & 2 & 3 & 4 & 5 & 6 & 7 & 8 & 9 & 10 & 11 & 12 & 13 & 14 & 15 & 16 \\
\hline
0.0 & 0.4672 & -- & -- & -- & -- & -- & -- & -- & -- & -- & -- & -- & -- & -- & -- & -- \\
0.1 & -- & 0.4600 & 0.4756 & 0.4760 & 0.4764 & 0.4756 & 0.4785 & 0.4764 & 0.4743 & 0.4689 & 0.4701 & 0.4739 & 0.4714 & 0.4663 & 0.4722 & 0.4853 \\
0.2 & -- & 0.4701 & 0.4680 & 0.4840 & 0.4672 & 0.4827 & 0.4705 & 0.4819 & 0.4790 & 0.4832 & 0.4886 & 0.4861 & 0.4811 & 0.4802 & 0.4891 & 0.4773 \\
0.3 & -- & 0.4655 & 0.4731 & 0.4781 & 0.4798 & 0.4764 & 0.4903 & 0.4983 & 0.4697 & 0.4916 & 0.4811 & 0.4920 & 0.4697 & 0.4844 & 0.4865 & 0.4903 \\
0.4 & -- & 0.4773 & 0.4689 & 0.4739 & 0.4802 & 0.4705 & 0.4815 & 0.4806 & 0.4844 & 0.4907 & 0.4676 & 0.4827 & 0.5000 & 0.4752 & 0.4815 & 0.4823 \\
0.5 & -- & 0.4781 & 0.4575 & 0.4769 & 0.4726 & 0.4840 & 0.4836 & 0.4979 & 0.4844 & 0.4899 & 0.4903 & 0.4912 & 0.4899 & 0.4823 & 0.4781 & -- \\
0.6 & -- & 0.4735 & 0.4722 & 0.4714 & 0.4726 & 0.4840 & 0.4865 & 0.5000 & 0.4827 & 0.4832 & 0.4769 & 0.4857 & 0.4954 & 0.4790 & 0.4899 & 0.4836 \\
0.7 & -- & 0.4684 & 0.4676 & 0.4705 & 0.4651 & 0.4823 & 0.4811 & 0.4941 & 0.4815 & 0.4886 & 0.4827 & 0.4907 & 0.4764 & 0.4865 & 0.4823 & 0.4756 \\
0.8 & -- & 0.4781 & 0.4646 & 0.4600 & 0.4689 & 0.4731 & 0.4735 & 0.5029 & 0.4735 & 0.4962 & 0.4790 & 0.4752 & 0.4697 & 0.4668 & 0.4832 & 0.4840 \\
0.9 & -- & 0.4676 & 0.4722 & 0.4562 & 0.4668 & 0.4651 & 0.4668 & 0.4819 & 0.4832 & 0.4827 & 0.4638 & 0.4646 & 0.4781 & 0.4693 & 0.4693 & 0.4621 \\
\hline
\end{tabular}%
}
\end{table}

\begin{table}[htbp]
\centering
\caption{Final BoolQ evals with varying $r$ and $s$, for 270M models trained for 20k total steps. See Table~\ref{tab:results} for training details.}
\label{tab:data}
\resizebox{\textwidth}{!}{%
\begin{tabular}{c|cccccccccccccccc}
\hline
\diagbox{$r$}{$s$} & 1 & 2 & 3 & 4 & 5 & 6 & 7 & 8 & 9 & 10 & 11 & 12 & 13 & 14 & 15 & 16 \\
\hline
0.0 & 0.5667 & -- & -- & -- & -- & -- & -- & -- & -- & -- & -- & -- & -- & -- & -- & -- \\
0.1 & -- & 0.6049 & 0.5893 & 0.5700 & 0.4606 & 0.5566 & 0.5734 & 0.5832 & 0.5532 & 0.6000 & 0.5208 & 0.5578 & 0.5141 & 0.5951 & 0.5927 & 0.5985 \\
0.2 & -- & 0.5911 & 0.5731 & 0.5502 & 0.5878 & 0.5391 & 0.5737 & 0.5982 & 0.5636 & 0.5994 & 0.5703 & 0.5865 & 0.5532 & 0.5456 & 0.5887 & 0.5630 \\
0.3 & -- & 0.5734 & 0.5884 & 0.5343 & 0.5575 & 0.5398 & 0.5930 & 0.5609 & 0.5624 & 0.5798 & 0.5752 & 0.5440 & 0.5881 & 0.5391 & 0.5856 & 0.5951 \\
0.4 & -- & 0.5379 & 0.5972 & 0.5242 & 0.5483 & 0.5190 & 0.5979 & 0.5538 & 0.5832 & 0.5792 & 0.5560 & 0.5700 & 0.5618 & 0.5765 & 0.5832 & 0.5841 \\
0.5 & -- & 0.5489 & 0.5902 & 0.5498 & 0.5780 & 0.5765 & 0.5817 & 0.5440 & 0.5437 & 0.5324 & 0.5618 & 0.5902 & 0.5578 & 0.5615 & 0.5948 & -- \\
0.6 & -- & 0.5462 & 0.5563 & 0.5884 & 0.5642 & 0.5303 & 0.5823 & 0.5679 & 0.5477 & 0.5489 & 0.5349 & 0.5468 & 0.5606 & 0.5862 & 0.5327 & 0.5862 \\
0.7 & -- & 0.5853 & 0.5862 & 0.5856 & 0.5761 & 0.5382 & 0.5618 & 0.5752 & 0.5609 & 0.5306 & 0.5474 & 0.5428 & 0.5740 & 0.5520 & 0.5657 & 0.5135 \\
0.8 & -- & 0.5850 & 0.5869 & 0.5813 & 0.5807 & 0.5471 & 0.5869 & 0.5336 & 0.5633 & 0.5474 & 0.5682 & 0.5404 & 0.5465 & 0.5945 & 0.5480 & 0.5795 \\
0.9 & -- & 0.5789 & 0.5835 & 0.5810 & 0.5523 & 0.5737 & 0.5829 & 0.5924 & 0.5884 & 0.5587 & 0.5725 & 0.5737 & 0.5664 & 0.5826 & 0.5869 & 0.5722 \\
\hline
\end{tabular}%
}
\end{table}

\begin{table}[htbp]
\centering
\caption{Final HellaSwag evals with varying $r$ and $s$, for 270M models trained for 20k total steps. See Table~\ref{tab:results} for training details.}
\label{tab:data}
\resizebox{\textwidth}{!}{%
\begin{tabular}{c|cccccccccccccccc}
\hline
\diagbox{$r$}{$s$} & 1 & 2 & 3 & 4 & 5 & 6 & 7 & 8 & 9 & 10 & 11 & 12 & 13 & 14 & 15 & 16 \\
\hline
0.0 & 0.3634 & -- & -- & -- & -- & -- & -- & -- & -- & -- & -- & -- & -- & -- & -- & -- \\
0.1 & -- & 0.3689 & 0.3766 & 0.3861 & 0.3832 & 0.3859 & 0.3865 & 0.3849 & 0.3858 & 0.3875 & 0.3841 & 0.3871 & 0.3853 & 0.3848 & 0.3852 & 0.3846 \\
0.2 & -- & 0.3775 & 0.3860 & 0.3905 & 0.3926 & 0.3900 & 0.3909 & 0.3931 & 0.3921 & 0.3881 & 0.3869 & 0.3909 & 0.3891 & 0.3886 & 0.3851 & 0.3900 \\
0.3 & -- & 0.3789 & 0.3806 & 0.3897 & 0.3935 & 0.3856 & 0.3899 & 0.3893 & 0.3898 & 0.3907 & 0.3869 & 0.3870 & 0.3902 & 0.3848 & 0.3873 & 0.3858 \\
0.4 & -- & 0.3756 & 0.3850 & 0.3914 & 0.3920 & 0.3892 & 0.3911 & 0.3911 & 0.3882 & 0.3872 & 0.3847 & 0.3903 & 0.3860 & 0.3922 & 0.3854 & 0.3857 \\
0.5 & -- & 0.3752 & 0.3847 & 0.3889 & 0.3902 & 0.3883 & 0.3923 & 0.3831 & 0.3884 & 0.3854 & 0.3819 & 0.3843 & 0.3826 & 0.3843 & 0.3844 & -- \\
0.6 & -- & 0.3765 & 0.3805 & 0.3876 & 0.3886 & 0.3822 & 0.3809 & 0.3845 & 0.3837 & 0.3807 & 0.3785 & 0.3832 & 0.3816 & 0.3828 & 0.3758 & 0.3787 \\
0.7 & -- & 0.3745 & 0.3833 & 0.3865 & 0.3844 & 0.3781 & 0.3774 & 0.3777 & 0.3790 & 0.3773 & 0.3746 & 0.3745 & 0.3713 & 0.3743 & 0.3712 & 0.3731 \\
0.8 & -- & 0.3690 & 0.3781 & 0.3818 & 0.3786 & 0.3706 & 0.3704 & 0.3702 & 0.3705 & 0.3687 & 0.3638 & 0.3614 & 0.3675 & 0.3617 & 0.3609 & 0.3593 \\
0.9 & -- & 0.3665 & 0.3681 & 0.3676 & 0.3615 & 0.3540 & 0.3575 & 0.3505 & 0.3510 & 0.3439 & 0.3422 & 0.3420 & 0.3347 & 0.3342 & 0.3354 & 0.3360 \\
\hline
\end{tabular}%
}
\end{table}

\begin{table}[htbp]
\centering
\caption{Final OpenBookQA evals with varying $r$ and $s$, for 270M models trained for 20k total steps. See Table~\ref{tab:results} for training details.}
\label{tab:data}
\resizebox{\textwidth}{!}{%
\begin{tabular}{c|cccccccccccccccc}
\hline
\diagbox{$r$}{$s$} & 1 & 2 & 3 & 4 & 5 & 6 & 7 & 8 & 9 & 10 & 11 & 12 & 13 & 14 & 15 & 16 \\
\hline
0.0 & 0.2920 & -- & -- & -- & -- & -- & -- & -- & -- & -- & -- & -- & -- & -- & -- & -- \\
0.1 & -- & 0.2960 & 0.2960 & 0.3160 & 0.3020 & 0.3060 & 0.3060 & 0.3160 & 0.3060 & 0.3000 & 0.3140 & 0.2960 & 0.3080 & 0.2940 & 0.3240 & 0.3120 \\
0.2 & -- & 0.3060 & 0.3060 & 0.2980 & 0.2820 & 0.2960 & 0.3160 & 0.3000 & 0.2940 & 0.3040 & 0.3120 & 0.3080 & 0.2920 & 0.3000 & 0.3060 & 0.3040 \\
0.3 & -- & 0.2900 & 0.3120 & 0.3000 & 0.3060 & 0.2980 & 0.3080 & 0.3060 & 0.3120 & 0.3160 & 0.3180 & 0.3060 & 0.3100 & 0.2860 & 0.3040 & 0.2940 \\
0.4 & -- & 0.3100 & 0.2940 & 0.3060 & 0.3040 & 0.3100 & 0.2900 & 0.2960 & 0.2940 & 0.3140 & 0.3040 & 0.2840 & 0.3020 & 0.3020 & 0.3220 & 0.2920 \\
0.5 & -- & 0.2980 & 0.3020 & 0.3160 & 0.2900 & 0.3240 & 0.2860 & 0.2940 & 0.2900 & 0.3180 & 0.2800 & 0.2920 & 0.3060 & 0.2980 & 0.3100 & -- \\
0.6 & -- & 0.3000 & 0.2860 & 0.3180 & 0.3180 & 0.3060 & 0.3040 & 0.2900 & 0.2880 & 0.3000 & 0.2940 & 0.2780 & 0.2880 & 0.2980 & 0.3140 & 0.2940 \\
0.7 & -- & 0.2980 & 0.2980 & 0.3100 & 0.3140 & 0.3060 & 0.2800 & 0.2880 & 0.2780 & 0.3060 & 0.2860 & 0.2920 & 0.2900 & 0.2960 & 0.2980 & 0.2900 \\
0.8 & -- & 0.2980 & 0.2780 & 0.3100 & 0.3100 & 0.3040 & 0.2860 & 0.2920 & 0.2740 & 0.2940 & 0.2940 & 0.2960 & 0.2940 & 0.2860 & 0.2980 & 0.2980 \\
0.9 & -- & 0.2980 & 0.2880 & 0.3100 & 0.3000 & 0.3040 & 0.2760 & 0.2800 & 0.2760 & 0.2860 & 0.3000 & 0.2960 & 0.2820 & 0.2900 & 0.3020 & 0.2880 \\
\hline
\end{tabular}%
}
\end{table}

\begin{table}[htbp]
\centering
\caption{Final PIQA evals with varying $r$ and $s$, for 270M models trained for 20k total steps. See Table~\ref{tab:results} for training details.}
\label{tab:data}
\resizebox{\textwidth}{!}{%
\begin{tabular}{c|cccccccccccccccc}
\hline
\diagbox{$r$}{$s$} & 1 & 2 & 3 & 4 & 5 & 6 & 7 & 8 & 9 & 10 & 11 & 12 & 13 & 14 & 15 & 16 \\
\hline
0.0 & 0.6638 & -- & -- & -- & -- & -- & -- & -- & -- & -- & -- & -- & -- & -- & -- & -- \\
0.1 & -- & 0.6697 & 0.6736 & 0.6719 & 0.6654 & 0.6643 & 0.6714 & 0.6752 & 0.6703 & 0.6817 & 0.6730 & 0.6725 & 0.6757 & 0.6681 & 0.6659 & 0.6736 \\
0.2 & -- & 0.6703 & 0.6708 & 0.6741 & 0.6659 & 0.6703 & 0.6714 & 0.6725 & 0.6703 & 0.6681 & 0.6643 & 0.6801 & 0.6659 & 0.6785 & 0.6801 & 0.6692 \\
0.3 & -- & 0.6676 & 0.6659 & 0.6665 & 0.6790 & 0.6703 & 0.6708 & 0.6708 & 0.6708 & 0.6708 & 0.6703 & 0.6757 & 0.6708 & 0.6779 & 0.6779 & 0.6616 \\
0.4 & -- & 0.6621 & 0.6654 & 0.6643 & 0.6627 & 0.6790 & 0.6719 & 0.6817 & 0.6681 & 0.6763 & 0.6757 & 0.6708 & 0.6746 & 0.6692 & 0.6752 & 0.6665 \\
0.5 & -- & 0.6708 & 0.6768 & 0.6578 & 0.6741 & 0.6763 & 0.6665 & 0.6659 & 0.6736 & 0.6736 & 0.6670 & 0.6736 & 0.6730 & 0.6752 & 0.6725 & -- \\
0.6 & -- & 0.6649 & 0.6676 & 0.6605 & 0.6665 & 0.6768 & 0.6714 & 0.6621 & 0.6741 & 0.6714 & 0.6703 & 0.6687 & 0.6681 & 0.6736 & 0.6741 & 0.6741 \\
0.7 & -- & 0.6649 & 0.6610 & 0.6632 & 0.6659 & 0.6719 & 0.6572 & 0.6714 & 0.6714 & 0.6752 & 0.6757 & 0.6638 & 0.6610 & 0.6801 & 0.6681 & 0.6567 \\
0.8 & -- & 0.6572 & 0.6676 & 0.6545 & 0.6621 & 0.6736 & 0.6654 & 0.6687 & 0.6621 & 0.6654 & 0.6703 & 0.6659 & 0.6578 & 0.6567 & 0.6649 & 0.6627 \\
0.9 & -- & 0.6610 & 0.6643 & 0.6513 & 0.6453 & 0.6600 & 0.6643 & 0.6502 & 0.6420 & 0.6534 & 0.6534 & 0.6442 & 0.6436 & 0.6507 & 0.6583 & 0.6534 \\
\hline
\end{tabular}%
}
\end{table}

\begin{table}[htbp]
\caption{Final Winogrande evals with varying $r$ and $s$, for 270M models trained for 20k total steps. See Table~\ref{tab:results} for training details.}
\label{tab:winogrande}
\centering
\resizebox{\textwidth}{!}{%
\begin{tabular}{c|cccccccccccccccc}
\hline
\diagbox{$r$}{$s$} & 1 & 2 & 3 & 4 & 5 & 6 & 7 & 8 & 9 & 10 & 11 & 12 & 13 & 14 & 15 & 16 \\
\hline
0.0 & 0.5130 & -- & -- & -- & -- & -- & -- & -- & -- & -- & -- & -- & -- & -- & -- & -- \\
0.1 & -- & 0.5193 & 0.5312 & 0.5107 & 0.5130 & 0.5257 & 0.5114 & 0.5193 & 0.5099 & 0.5162 & 0.5375 & 0.5209 & 0.5264 & 0.5091 & 0.5154 & 0.5241 \\
0.2 & -- & 0.5114 & 0.4957 & 0.5185 & 0.5146 & 0.5264 & 0.5091 & 0.5391 & 0.5114 & 0.5067 & 0.5185 & 0.5162 & 0.5249 & 0.4988 & 0.5257 & 0.5257 \\
0.3 & -- & 0.5367 & 0.5091 & 0.5233 & 0.5233 & 0.5107 & 0.5020 & 0.5083 & 0.5280 & 0.5099 & 0.5280 & 0.5028 & 0.4957 & 0.5225 & 0.5107 & 0.5446 \\
0.4 & -- & 0.5130 & 0.5280 & 0.5130 & 0.5233 & 0.5130 & 0.5099 & 0.5241 & 0.5028 & 0.5217 & 0.5099 & 0.5272 & 0.5178 & 0.4925 & 0.5185 & 0.5028 \\
0.5 & -- & 0.5414 & 0.5154 & 0.5154 & 0.5138 & 0.5257 & 0.5367 & 0.5233 & 0.5028 & 0.5067 & 0.5178 & 0.5328 & 0.5146 & 0.5154 & 0.5391 & -- \\
0.6 & -- & 0.5146 & 0.5067 & 0.5146 & 0.5162 & 0.5036 & 0.5059 & 0.5296 & 0.5075 & 0.5083 & 0.5162 & 0.5020 & 0.5107 & 0.5264 & 0.5233 & 0.5146 \\
0.7 & -- & 0.5170 & 0.5099 & 0.4988 & 0.5004 & 0.5178 & 0.5083 & 0.5059 & 0.5217 & 0.5043 & 0.5091 & 0.5280 & 0.5004 & 0.4988 & 0.5351 & 0.5051 \\
0.8 & -- & 0.5233 & 0.5051 & 0.4901 & 0.5114 & 0.5028 & 0.5178 & 0.5185 & 0.5146 & 0.5178 & 0.5170 & 0.5075 & 0.5091 & 0.5099 & 0.5012 & 0.5114 \\
0.9 & -- & 0.5209 & 0.5020 & 0.5028 & 0.5020 & 0.4972 & 0.5067 & 0.5130 & 0.5162 & 0.5146 & 0.5114 & 0.5114 & 0.4925 & 0.5249 & 0.5020 & 0.5020 \\
\hline
\end{tabular}%
}
\end{table}

\begin{table}[htbp]
\caption{Final ARC-Challenge evals with varying $r$ and $s$, for 270M models trained for 100k total steps. See Table~\ref{tab:results} for training details.}
\label{tab:arc_challenge_270m}
\centering
\resizebox{\textwidth}{!}{%
\begin{tabular}{c|ccccccccccccc}
\hline
\diagbox{$r$}{$s$} & 1 & 2 & 3 & 4 & 5 & 6 & 7 & 8 & 9 & 10 & 11 & 12 & 13 \\
\hline
0.0 & 0.2619 & -- & -- & -- & -- & -- & -- & -- & -- & -- & -- & -- & -- \\
0.1 & -- & 0.2568 & 0.2619 & 0.2534 & 0.2671 & 0.2654 & 0.2654 & 0.2654 & 0.2688 & 0.2645 & 0.2440 & 0.2543 & 0.2782 \\
0.2 & -- & 0.2517 & 0.2654 & 0.2662 & 0.2637 & 0.2645 & 0.2577 & 0.2619 & 0.2628 & 0.2773 & 0.2594 & 0.2730 & 0.2756 \\
0.3 & -- & 0.2611 & 0.2577 & 0.2705 & 0.2568 & 0.2551 & 0.2534 & 0.2696 & 0.2602 & 0.2662 & 0.2739 & 0.2688 & 0.2645 \\
0.4 & -- & 0.2628 & 0.2568 & 0.2517 & 0.2568 & 0.2611 & 0.2722 & 0.2739 & 0.2662 & 0.2654 & 0.2585 & 0.2611 & 0.2611 \\
0.5 & -- & 0.2449 & 0.2585 & 0.2577 & 0.2662 & 0.2645 & 0.2628 & 0.2602 & 0.2637 & 0.2602 & 0.2534 & 0.2628 & 0.2688 \\
0.6 & -- & 0.2500 & 0.2602 & 0.2543 & 0.2585 & 0.2577 & 0.2577 & 0.2611 & 0.2602 & 0.2773 & 0.2739 & 0.2611 & 0.2560 \\
0.7 & -- & 0.2534 & 0.2705 & 0.2637 & 0.2560 & 0.2662 & 0.2705 & 0.2637 & 0.2645 & 0.2628 & 0.2705 & 0.2713 & 0.2679 \\
0.8 & -- & 0.2432 & 0.2611 & 0.2619 & 0.2432 & 0.2628 & 0.2594 & 0.2645 & 0.2483 & 0.2662 & 0.2619 & 0.2722 & 0.2602 \\
0.9 & -- & 0.2551 & 0.2602 & 0.2747 & 0.2483 & 0.2517 & 0.2474 & 0.2491 & 0.2747 & 0.2585 & 0.2466 & 0.2611 & 0.2594 \\
\hline
\end{tabular}%
}
\end{table}

\begin{table}[htbp]
\caption{Final ARC-Easy evals with varying $r$ and $s$, for 270M models trained for 100k total steps. See Table~\ref{tab:results} for training details.}
\label{tab:arc_easy}
\centering
\resizebox{\textwidth}{!}{%
\begin{tabular}{c|ccccccccccccc}
\hline
\diagbox{$r$}{$s$} & 1 & 2 & 3 & 4 & 5 & 6 & 7 & 8 & 9 & 10 & 11 & 12 & 13 \\
\hline
0.0 & 0.4747 & -- & -- & -- & -- & -- & -- & -- & -- & -- & -- & -- & -- \\
0.1 & -- & 0.4870 & 0.4941 & 0.4903 & 0.5029 & 0.5202 & 0.4954 & 0.5004 & 0.5198 & 0.5114 & 0.4992 & 0.4920 & 0.5088 \\
0.2 & -- & 0.4731 & 0.4924 & 0.4907 & 0.5332 & 0.5008 & 0.4895 & 0.4983 & 0.5130 & 0.4992 & 0.5051 & 0.5223 & 0.5034 \\
0.3 & -- & 0.4815 & 0.4979 & 0.5021 & 0.5160 & 0.5034 & 0.5181 & 0.5278 & 0.5173 & 0.5067 & 0.4945 & 0.5156 & 0.5114 \\
0.4 & -- & 0.4794 & 0.5025 & 0.4899 & 0.5051 & 0.5072 & 0.5017 & 0.5303 & 0.5223 & 0.5236 & 0.5181 & 0.5147 & 0.5021 \\
0.5 & -- & 0.4693 & 0.5038 & 0.5013 & 0.5189 & 0.4992 & 0.5156 & 0.5114 & 0.5072 & 0.5029 & 0.5177 & 0.5046 & 0.4983 \\
0.6 & -- & 0.4769 & 0.5093 & 0.5093 & 0.4920 & 0.5076 & 0.4962 & 0.5004 & 0.5059 & 0.4979 & 0.5004 & 0.5101 & 0.5114 \\
0.7 & -- & 0.4735 & 0.4933 & 0.4992 & 0.4832 & 0.4954 & 0.4987 & 0.4895 & 0.5000 & 0.5114 & 0.5122 & 0.4979 & 0.4827 \\
0.8 & -- & 0.4663 & 0.4836 & 0.4975 & 0.4886 & 0.4920 & 0.5067 & 0.5101 & 0.5042 & 0.4962 & 0.5059 & 0.5105 & 0.4987 \\
0.9 & -- & 0.4760 & 0.4920 & 0.4907 & 0.4823 & 0.4823 & 0.5021 & 0.5076 & 0.4996 & 0.4794 & 0.4815 & 0.5059 & 0.4853 \\
\hline
\end{tabular}%
}
\end{table}

\begin{table}[htbp]
\caption{Final BoolQ evals with varying $r$ and $s$, for 270M models trained for 100k total steps. See Table~\ref{tab:results} for training details.}
\label{tab:boolq_270m_100k}
\centering
\resizebox{\textwidth}{!}{%
\begin{tabular}{c|ccccccccccccc}
\hline
\diagbox{$r$}{$s$} & 1 & 2 & 3 & 4 & 5 & 6 & 7 & 8 & 9 & 10 & 11 & 12 & 13 \\
\hline
0.0 & 0.5847 & -- & -- & -- & -- & -- & -- & -- & -- & -- & -- & -- & -- \\
0.1 & -- & 0.5223 & 0.5606 & 0.5245 & 0.5713 & 0.5618 & 0.5076 & 0.5933 & 0.5572 & 0.5621 & 0.5336 & 0.5550 & 0.5786 \\
0.2 & -- & 0.5260 & 0.6000 & 0.5318 & 0.5960 & 0.5431 & 0.5798 & 0.5462 & 0.5621 & 0.5083 & 0.5713 & 0.5746 & 0.5309 \\
0.3 & -- & 0.5391 & 0.5841 & 0.5535 & 0.5410 & 0.5789 & 0.5761 & 0.5566 & 0.5847 & 0.4927 & 0.5352 & 0.5190 & 0.5673 \\
0.4 & -- & 0.5131 & 0.5713 & 0.5575 & 0.5557 & 0.5187 & 0.5685 & 0.5557 & 0.5336 & 0.5382 & 0.5272 & 0.4838 & 0.5872 \\
0.5 & -- & 0.6000 & 0.5046 & 0.5612 & 0.5468 & 0.5700 & 0.5596 & 0.5676 & 0.5930 & 0.5554 & 0.5685 & 0.5468 & 0.5789 \\
0.6 & -- & 0.6141 & 0.5092 & 0.5083 & 0.5979 & 0.5957 & 0.5988 & 0.5416 & 0.5300 & 0.5297 & 0.5339 & 0.5367 & 0.5633 \\
0.7 & -- & 0.5966 & 0.5306 & 0.5113 & 0.5480 & 0.5899 & 0.5471 & 0.5694 & 0.5346 & 0.5080 & 0.5119 & 0.4957 & 0.5318 \\
0.8 & -- & 0.5869 & 0.5495 & 0.4936 & 0.5615 & 0.5511 & 0.5859 & 0.5544 & 0.5621 & 0.4966 & 0.4547 & 0.5220 & 0.5554 \\
0.9 & -- & 0.6162 & 0.5431 & 0.5024 & 0.5832 & 0.5388 & 0.5719 & 0.5758 & 0.5709 & 0.5116 & 0.5554 & 0.5462 & 0.5560 \\
\hline
\end{tabular}%
}
\end{table}

\begin{table}[htbp]
\caption{Final HellaSwag evals with varying $r$ and $s$, for 270M models trained for 100k total steps. See Table~\ref{tab:results} for training details.}
\label{tab:boolq_270m}
\centering
\resizebox{\textwidth}{!}{%
\begin{tabular}{c|ccccccccccccc}
\hline
\diagbox{$r$}{$s$} & 1 & 2 & 3 & 4 & 5 & 6 & 7 & 8 & 9 & 10 & 11 & 12 & 13 \\
\hline
0.0 & 0.4021 & -- & -- & -- & -- & -- & -- & -- & -- & -- & -- & -- & -- \\
0.1 & -- & 0.4107 & 0.4190 & 0.4237 & 0.4225 & 0.4252 & 0.4284 & 0.4281 & 0.4261 & 0.4276 & 0.4227 & 0.4292 & 0.4309 \\
0.2 & -- & 0.4122 & 0.4176 & 0.4260 & 0.4277 & 0.4266 & 0.4269 & 0.4267 & 0.4267 & 0.4262 & 0.4288 & 0.4243 & 0.4280 \\
0.3 & -- & 0.4143 & 0.4179 & 0.4306 & 0.4275 & 0.4261 & 0.4260 & 0.4265 & 0.4284 & 0.4238 & 0.4275 & 0.4243 & 0.4236 \\
0.4 & -- & 0.4137 & 0.4200 & 0.4265 & 0.4340 & 0.4279 & 0.4248 & 0.4290 & 0.4266 & 0.4261 & 0.4291 & 0.4246 & 0.4247 \\
0.5 & -- & 0.4145 & 0.4227 & 0.4247 & 0.4272 & 0.4246 & 0.4282 & 0.4219 & 0.4251 & 0.4219 & 0.4256 & 0.4219 & 0.4266 \\
0.6 & -- & 0.4144 & 0.4192 & 0.4226 & 0.4250 & 0.4244 & 0.4231 & 0.4256 & 0.4262 & 0.4200 & 0.4196 & 0.4180 & 0.4180 \\
0.7 & -- & 0.4124 & 0.4192 & 0.4214 & 0.4229 & 0.4195 & 0.4206 & 0.4188 & 0.4220 & 0.4141 & 0.4164 & 0.4160 & 0.4148 \\
0.8 & -- & 0.4102 & 0.4164 & 0.4224 & 0.4152 & 0.4153 & 0.4171 & 0.4175 & 0.4132 & 0.4120 & 0.4113 & 0.4068 & 0.4077 \\
0.9 & -- & 0.4100 & 0.4091 & 0.4089 & 0.4097 & 0.4019 & 0.4020 & 0.4049 & 0.3976 & 0.3950 & 0.3940 & 0.3921 & 0.3919 \\
\hline
\end{tabular}%
}
\end{table}

\begin{table}[htbp]
\centering
\caption{Final OpenBookQA evals with varying $r$ and $s$, for 270M models trained for 100k total steps. See Table~\ref{tab:results} for training details.}
\label{tab:boolq}
\resizebox{\textwidth}{!}{%
\begin{tabular}{c|ccccccccccccc}
\hline
\diagbox{$r$}{$s$} & 1 & 2 & 3 & 4 & 5 & 6 & 7 & 8 & 9 & 10 & 11 & 12 & 13 \\
\hline
0.0 & 0.3060 & -- & -- & -- & -- & -- & -- & -- & -- & -- & -- & -- & -- \\
0.1 & -- & 0.3200 & 0.3020 & 0.3120 & 0.3100 & 0.3180 & 0.3080 & 0.2980 & 0.3140 & 0.3140 & 0.3140 & 0.3200 & 0.3340 \\
0.2 & -- & 0.3040 & 0.3220 & 0.3180 & 0.3220 & 0.3220 & 0.3020 & 0.3040 & 0.3100 & 0.3180 & 0.3220 & 0.3040 & 0.3240 \\
0.3 & -- & 0.3260 & 0.3400 & 0.3120 & 0.3260 & 0.3240 & 0.3220 & 0.3220 & 0.3180 & 0.3300 & 0.3060 & 0.3160 & 0.3160 \\
0.4 & -- & 0.2980 & 0.3100 & 0.3000 & 0.3220 & 0.3100 & 0.3080 & 0.3060 & 0.3200 & 0.3280 & 0.3040 & 0.3080 & 0.3040 \\
0.5 & -- & 0.3200 & 0.3160 & 0.3180 & 0.3240 & 0.3060 & 0.3220 & 0.3180 & 0.3000 & 0.3200 & 0.3140 & 0.3160 & 0.3260 \\
0.6 & -- & 0.3300 & 0.3300 & 0.3240 & 0.3040 & 0.3080 & 0.3120 & 0.3180 & 0.2980 & 0.3060 & 0.3080 & 0.3160 & 0.3240 \\
0.7 & -- & 0.2980 & 0.3080 & 0.3080 & 0.3180 & 0.3100 & 0.2900 & 0.3040 & 0.3060 & 0.2940 & 0.3080 & 0.3340 & 0.3020 \\
0.8 & -- & 0.3060 & 0.3160 & 0.3080 & 0.3220 & 0.3060 & 0.3120 & 0.2980 & 0.3200 & 0.2860 & 0.3320 & 0.2920 & 0.3020 \\
0.9 & -- & 0.2860 & 0.3280 & 0.3100 & 0.3180 & 0.3200 & 0.3060 & 0.3120 & 0.3140 & 0.2980 & 0.3080 & 0.2960 & 0.3020 \\
\hline
\end{tabular}%
}
\end{table}

\begin{table}[htbp]
\centering
\caption{Final PIQA evals with varying $r$ and $s$, for 270M models trained for 100k total steps. See Table~\ref{tab:results} for training details.}
\label{tab:boolq}
\resizebox{\textwidth}{!}{%
\begin{tabular}{c|ccccccccccccc}
\hline
\diagbox{$r$}{$s$} & 1 & 2 & 3 & 4 & 5 & 6 & 7 & 8 & 9 & 10 & 11 & 12 & 13 \\
\hline
0.0 & 0.6730 & -- & -- & -- & -- & -- & -- & -- & -- & -- & -- & -- & -- \\
0.1 & -- & 0.6746 & 0.6844 & 0.6785 & 0.6785 & 0.6942 & 0.6855 & 0.6861 & 0.6774 & 0.6834 & 0.6861 & 0.6812 & 0.6904 \\
0.2 & -- & 0.6812 & 0.6779 & 0.6801 & 0.6779 & 0.6921 & 0.6861 & 0.6899 & 0.6850 & 0.6844 & 0.6844 & 0.6861 & 0.6801 \\
0.3 & -- & 0.6785 & 0.6817 & 0.6844 & 0.6801 & 0.6904 & 0.6752 & 0.6790 & 0.6855 & 0.6823 & 0.6877 & 0.6893 & 0.6910 \\
0.4 & -- & 0.6790 & 0.6839 & 0.6828 & 0.6823 & 0.6872 & 0.6779 & 0.6839 & 0.6937 & 0.6823 & 0.6964 & 0.6866 & 0.6861 \\
0.5 & -- & 0.6844 & 0.6779 & 0.6785 & 0.6801 & 0.6937 & 0.6855 & 0.6855 & 0.6921 & 0.6844 & 0.6915 & 0.6877 & 0.6899 \\
0.6 & -- & 0.6806 & 0.6834 & 0.6834 & 0.6882 & 0.6899 & 0.6834 & 0.6948 & 0.6888 & 0.6834 & 0.6823 & 0.6839 & 0.6828 \\
0.7 & -- & 0.6708 & 0.6752 & 0.6757 & 0.6779 & 0.6926 & 0.6763 & 0.6921 & 0.6866 & 0.6823 & 0.6806 & 0.6806 & 0.6741 \\
0.8 & -- & 0.6665 & 0.6752 & 0.6779 & 0.6817 & 0.6844 & 0.6855 & 0.6806 & 0.6823 & 0.6844 & 0.6768 & 0.6828 & 0.6795 \\
0.9 & -- & 0.6665 & 0.6654 & 0.6692 & 0.6844 & 0.6806 & 0.6779 & 0.6861 & 0.6757 & 0.6741 & 0.6714 & 0.6681 & 0.6659 \\
\hline
\end{tabular}%
}
\end{table}

\begin{table}[htbp]
\centering
\caption{Final Winogrande evals with varying $r$ and $s$, for 270M models trained for 100k total steps. See Table~\ref{tab:results} for training details.}
\label{tab:winogrande}
\resizebox{\textwidth}{!}{%
\begin{tabular}{c|ccccccccccccc}
\hline
\diagbox{$r$}{$s$} & 1 & 2 & 3 & 4 & 5 & 6 & 7 & 8 & 9 & 10 & 11 & 12 & 13 \\
\hline
0.0 & 0.5154 & -- & -- & -- & -- & -- & -- & -- & -- & -- & -- & -- & -- \\
0.1 & -- & 0.5170 & 0.5351 & 0.5036 & 0.5209 & 0.5343 & 0.5272 & 0.5272 & 0.5351 & 0.5328 & 0.5391 & 0.5446 & 0.5122 \\
0.2 & -- & 0.5193 & 0.5288 & 0.5304 & 0.5406 & 0.5454 & 0.5422 & 0.5272 & 0.5399 & 0.5288 & 0.5383 & 0.5193 & 0.5343 \\
0.3 & -- & 0.5185 & 0.5170 & 0.5320 & 0.5257 & 0.5422 & 0.5296 & 0.5414 & 0.5193 & 0.5304 & 0.5296 & 0.5375 & 0.5288 \\
0.4 & -- & 0.5288 & 0.5280 & 0.5201 & 0.5414 & 0.5257 & 0.5406 & 0.5383 & 0.5470 & 0.5201 & 0.5288 & 0.5375 & 0.5280 \\
0.5 & -- & 0.5075 & 0.5146 & 0.5335 & 0.5304 & 0.5320 & 0.5430 & 0.5320 & 0.5304 & 0.5162 & 0.5343 & 0.5264 & 0.5438 \\
0.6 & -- & 0.4957 & 0.5296 & 0.5241 & 0.5312 & 0.5383 & 0.5296 & 0.5375 & 0.5383 & 0.5288 & 0.5091 & 0.5462 & 0.5414 \\
0.7 & -- & 0.5130 & 0.5225 & 0.5375 & 0.5249 & 0.5264 & 0.5320 & 0.5391 & 0.5264 & 0.5225 & 0.5280 & 0.5296 & 0.5209 \\
0.8 & -- & 0.5233 & 0.5257 & 0.5083 & 0.5051 & 0.5430 & 0.5272 & 0.5272 & 0.5257 & 0.5304 & 0.5280 & 0.5359 & 0.5114 \\
0.9 & -- & 0.5280 & 0.5178 & 0.5170 & 0.5193 & 0.5178 & 0.5170 & 0.5083 & 0.5178 & 0.5075 & 0.5209 & 0.5067 & 0.5178 \\
\hline
\end{tabular}%
}
\end{table}

\begin{table}[htbp]
\caption{Final ARC-Challenge evals with varying $r$ and $s$, for 600M models trained for 20k total steps. See Table~\ref{tab:results} for training details.}
\label{tab:arc_challenge}
\centering
\resizebox{\textwidth}{!}{%
\begin{tabular}{c|ccccccccccccc}
\hline
\diagbox{$r$}{$s$} & 1 & 2 & 3 & 4 & 5 & 6 & 7 & 8 & 9 & 10 & 11 & 12 & 13 \\
\hline
0.0 & 0.2551 & -- & -- & -- & -- & -- & -- & -- & -- & -- & -- & -- & -- \\
0.1 & -- & 0.2628 & 0.2671 & 0.2816 & 0.2747 & 0.2833 & 0.2952 & 0.2739 & 0.2892 & 0.2850 & 0.2765 & 0.2833 & 0.2790 \\
0.2 & -- & 0.2679 & 0.2765 & 0.2782 & 0.2782 & 0.2790 & 0.2867 & 0.3029 & 0.2867 & 0.2799 & 0.2918 & 0.3012 & 0.2927 \\
0.3 & -- & 0.2645 & 0.2765 & 0.2688 & 0.2705 & 0.2688 & 0.2944 & 0.2824 & 0.2824 & 0.2816 & 0.2961 & 0.2969 & 0.2816 \\
0.4 & -- & 0.2628 & 0.2782 & 0.2705 & 0.2850 & 0.2730 & 0.2867 & 0.2875 & 0.2910 & 0.2901 & 0.2850 & 0.2892 & 0.3063 \\
0.5 & -- & 0.2765 & 0.2807 & 0.2884 & 0.2782 & 0.2611 & 0.2961 & 0.2824 & 0.2910 & 0.2961 & 0.2807 & 0.3012 & 0.2952 \\
0.6 & -- & 0.2688 & 0.2696 & 0.2799 & 0.2773 & 0.2671 & 0.2910 & 0.2765 & 0.2858 & 0.2935 & 0.2824 & 0.2995 & 0.2892 \\
0.7 & -- & 0.2696 & 0.2688 & 0.2867 & 0.2688 & 0.2782 & 0.2816 & 0.2858 & 0.2816 & 0.2875 & 0.2961 & 0.2910 & 0.2918 \\
0.8 & -- & 0.2747 & 0.2875 & 0.2944 & 0.2799 & 0.2696 & 0.2637 & 0.2824 & 0.2884 & 0.2875 & 0.2867 & 0.2816 & 0.2782 \\
0.9 & -- & 0.2645 & 0.2722 & 0.2671 & 0.2816 & 0.2688 & 0.2628 & 0.2594 & 0.2705 & 0.2705 & 0.2773 & 0.2747 & 0.2730 \\
\hline
\end{tabular}%
}
\end{table}

\begin{table}[htbp]
\caption{Final ARC-Easy evals with varying $r$ and $s$, for 600M models trained for 20k total steps. See Table~\ref{tab:results} for training details.}
\label{tab:arc_easy}
\centering
\resizebox{\textwidth}{!}{%
\begin{tabular}{c|ccccccccccccc}
\hline
\diagbox{$r$}{$s$} & 1 & 2 & 3 & 4 & 5 & 6 & 7 & 8 & 9 & 10 & 11 & 12 & 13 \\
\hline
0.0 & 0.5168 & - & - & - & - & - & - & - & - & - & - & - & - \\
0.1 & - & 0.5210 & 0.5530 & 0.5412 & 0.5307 & 0.5320 & 0.5286 & 0.5391 & 0.5219 & 0.5383 & 0.5370 & 0.5480 & 0.5526 \\
0.2 & - & 0.5261 & 0.5328 & 0.5543 & 0.5429 & 0.5202 & 0.5362 & 0.5442 & 0.5501 & 0.5450 & 0.5644 & 0.5564 & 0.5593 \\
0.3 & - & 0.5152 & 0.5391 & 0.5450 & 0.5261 & 0.5248 & 0.5383 & 0.5320 & 0.5543 & 0.5535 & 0.5492 & 0.5606 & 0.5455 \\
0.4 & - & 0.5173 & 0.5438 & 0.5610 & 0.5265 & 0.5236 & 0.5379 & 0.5467 & 0.5400 & 0.5522 & 0.5467 & 0.5619 & 0.5556 \\
0.5 & - & 0.5118 & 0.5349 & 0.5484 & 0.5362 & 0.5248 & 0.5442 & 0.5299 & 0.5547 & 0.5450 & 0.5425 & 0.5661 & 0.5593 \\
0.6 & - & 0.5038 & 0.5459 & 0.5412 & 0.5455 & 0.5417 & 0.5467 & 0.5417 & 0.5543 & 0.5665 & 0.5467 & 0.5657 & 0.5518 \\
0.7 & - & 0.5093 & 0.5366 & 0.5396 & 0.5425 & 0.5379 & 0.5539 & 0.5311 & 0.5614 & 0.5459 & 0.5421 & 0.5530 & 0.5366 \\
0.8 & - & 0.5093 & 0.5400 & 0.5446 & 0.5375 & 0.5332 & 0.5463 & 0.5370 & 0.5341 & 0.5442 & 0.5497 & 0.5450 & 0.5332 \\
0.9 & - & 0.5000 & 0.5370 & 0.5311 & 0.5236 & 0.5311 & 0.5290 & 0.5215 & 0.5223 & 0.5404 & 0.5265 & 0.5408 & 0.5434 \\
\hline
\end{tabular}%
}
\end{table}

\begin{table}[htbp]
\centering
\caption{Final BoolQ evals with varying $r$ and $s$, for 600M models trained for 20k total steps. See Table~\ref{tab:results} for training details.}
\label{tab:boolq}
\resizebox{\textwidth}{!}{%
\begin{tabular}{c|ccccccccccccc}
\hline
\diagbox{$r$}{$s$} & 1 & 2 & 3 & 4 & 5 & 6 & 7 & 8 & 9 & 10 & 11 & 12 & 13 \\
\hline
0.0 & 0.5661 & -- & -- & -- & -- & -- & -- & -- & -- & -- & -- & -- & -- \\
0.1 & -- & 0.5924 & 0.5924 & 0.5896 & 0.6098 & 0.5517 & 0.5920 & 0.6214 & 0.6031 & 0.5722 & 0.5697 & 0.5642 & 0.5789 \\
0.2 & -- & 0.5761 & 0.5991 & 0.5994 & 0.6254 & 0.6070 & 0.5783 & 0.5612 & 0.5364 & 0.5300 & 0.5327 & 0.6110 & 0.6012 \\
0.3 & -- & 0.6000 & 0.5810 & 0.5988 & 0.6024 & 0.5477 & 0.5336 & 0.5333 & 0.5483 & 0.5462 & 0.5456 & 0.5865 & 0.5670 \\
0.4 & -- & 0.5661 & 0.5872 & 0.5963 & 0.6037 & 0.5703 & 0.5297 & 0.5664 & 0.5703 & 0.5550 & 0.5532 & 0.5768 & 0.5911 \\
0.5 & -- & 0.5832 & 0.6009 & 0.6052 & 0.6162 & 0.5615 & 0.5697 & 0.5471 & 0.5070 & 0.5673 & 0.5119 & 0.5330 & 0.5985 \\
0.6 & -- & 0.5394 & 0.5823 & 0.6257 & 0.5942 & 0.5899 & 0.6024 & 0.5520 & 0.5654 & 0.5410 & 0.5346 & 0.5453 & 0.5664 \\
0.7 & -- & 0.5798 & 0.5627 & 0.6110 & 0.6043 & 0.5847 & 0.5624 & 0.5780 & 0.5777 & 0.5587 & 0.5125 & 0.6135 & 0.5884 \\
0.8 & -- & 0.5813 & 0.5838 & 0.6187 & 0.6168 & 0.5960 & 0.5758 & 0.5636 & 0.5847 & 0.5453 & 0.5211 & 0.5905 & 0.5755 \\
0.9 & -- & 0.5761 & 0.5869 & 0.5960 & 0.6183 & 0.5645 & 0.5355 & 0.5498 & 0.5905 & 0.5829 & 0.5951 & 0.6110 & 0.5648 \\
\hline
\end{tabular}%
}
\end{table}

\begin{table}[htbp]
\centering
\caption{Final HellaSwag evals with varying $r$ and $s$, for 600M models trained for 20k total steps. See Table~\ref{tab:results} for training details.}
\label{tab:hellaswag}
\resizebox{\textwidth}{!}{%
\begin{tabular}{c|ccccccccccccc}
\hline
\diagbox{$r$}{$s$} & 1 & 2 & 3 & 4 & 5 & 6 & 7 & 8 & 9 & 10 & 11 & 12 & 13 \\
\hline
0.0 & 0.4347 & -- & -- & -- & -- & -- & -- & -- & -- & -- & -- & -- & -- \\
0.1 & -- & 0.4554 & 0.4658 & 0.4780 & 0.4763 & 0.4736 & 0.4760 & 0.4756 & 0.4750 & 0.4771 & 0.4739 & 0.4787 & 0.4764 \\
0.2 & -- & 0.4628 & 0.4711 & 0.4806 & 0.4802 & 0.4810 & 0.4823 & 0.4855 & 0.4832 & 0.4813 & 0.4797 & 0.4809 & 0.4811 \\
0.3 & -- & 0.4620 & 0.4726 & 0.4788 & 0.4832 & 0.4817 & 0.4777 & 0.4840 & 0.4837 & 0.4825 & 0.4825 & 0.4793 & 0.4806 \\
0.4 & -- & 0.4613 & 0.4763 & 0.4797 & 0.4814 & 0.4802 & 0.4804 & 0.4812 & 0.4811 & 0.4792 & 0.4727 & 0.4749 & 0.4753 \\
0.5 & -- & 0.4642 & 0.4783 & 0.4812 & 0.4808 & 0.4781 & 0.4795 & 0.4767 & 0.4795 & 0.4757 & 0.4810 & 0.4718 & 0.4725 \\
0.6 & -- & 0.4614 & 0.4710 & 0.4780 & 0.4754 & 0.4755 & 0.4717 & 0.4695 & 0.4721 & 0.4706 & 0.4689 & 0.4671 & 0.4631 \\
0.7 & -- & 0.4588 & 0.4682 & 0.4740 & 0.4702 & 0.4706 & 0.4648 & 0.4679 & 0.4654 & 0.4590 & 0.4593 & 0.4556 & 0.4527 \\
0.8 & -- & 0.4561 & 0.4658 & 0.4685 & 0.4588 & 0.4582 & 0.4568 & 0.4524 & 0.4485 & 0.4465 & 0.4448 & 0.4414 & 0.4367 \\
0.9 & -- & 0.4494 & 0.4507 & 0.4482 & 0.4417 & 0.4343 & 0.4265 & 0.4257 & 0.4207 & 0.4107 & 0.4076 & 0.4033 & 0.4005 \\
\hline
\end{tabular}%
}
\end{table}

\begin{table}[htbp]
\centering
\caption{Final OpenBookQA evals with varying $r$ and $s$, for 600M models trained for 20k total steps. See Table~\ref{tab:results} for training details.}
\label{tab:hellaswag}
\resizebox{\textwidth}{!}{%
\begin{tabular}{c|ccccccccccccc}
\hline
\diagbox{$r$}{$s$} & 1 & 2 & 3 & 4 & 5 & 6 & 7 & 8 & 9 & 10 & 11 & 12 & 13 \\
\hline
0.0 & 0.3120 & -- & -- & -- & -- & -- & -- & -- & -- & -- & -- & -- & -- \\
0.1 & -- & 0.3200 & 0.3220 & 0.3280 & 0.3440 & 0.3460 & 0.3320 & 0.3480 & 0.3160 & 0.3540 & 0.3560 & 0.3220 & 0.3480 \\
0.2 & -- & 0.3360 & 0.3260 & 0.3320 & 0.3380 & 0.3160 & 0.3300 & 0.3220 & 0.3280 & 0.3360 & 0.3260 & 0.3340 & 0.3380 \\
0.3 & -- & 0.3280 & 0.3340 & 0.3280 & 0.3100 & 0.3500 & 0.3340 & 0.3480 & 0.3340 & 0.3420 & 0.3240 & 0.3400 & 0.3180 \\
0.4 & -- & 0.3320 & 0.3340 & 0.3440 & 0.3320 & 0.3460 & 0.3140 & 0.3220 & 0.3340 & 0.3300 & 0.3400 & 0.3380 & 0.3420 \\
0.5 & -- & 0.3320 & 0.3420 & 0.3180 & 0.3300 & 0.3280 & 0.3260 & 0.3360 & 0.3420 & 0.3060 & 0.3220 & 0.3220 & 0.3200 \\
0.6 & -- & 0.3320 & 0.3340 & 0.3320 & 0.3200 & 0.3320 & 0.3260 & 0.3240 & 0.3500 & 0.3200 & 0.3260 & 0.3460 & 0.3320 \\
0.7 & -- & 0.3380 & 0.3380 & 0.3480 & 0.3000 & 0.3380 & 0.3300 & 0.3500 & 0.3500 & 0.3140 & 0.3320 & 0.3400 & 0.3240 \\
0.8 & -- & 0.3260 & 0.3340 & 0.3240 & 0.3180 & 0.3160 & 0.3200 & 0.3400 & 0.3480 & 0.3380 & 0.3380 & 0.3240 & 0.3200 \\
0.9 & -- & 0.3040 & 0.3280 & 0.3140 & 0.3060 & 0.3080 & 0.3160 & 0.3280 & 0.3440 & 0.3180 & 0.3340 & 0.3060 & 0.3140 \\
\hline
\end{tabular}%
}
\end{table}

\begin{table}[htbp]
\caption{Final PIQA evals with varying $r$ and $s$, for 600M models trained for 20k total steps. See Table~\ref{tab:results} for training details.}
\label{tab:piqa}
\centering
\resizebox{\textwidth}{!}{%
\begin{tabular}{c|ccccccccccccc}
\hline
\diagbox{$r$}{$s$} & 1 & 2 & 3 & 4 & 5 & 6 & 7 & 8 & 9 & 10 & 11 & 12 & 13 \\
\hline
0.0 & 0.6904 & - & - & - & - & - & - & - & - & - & - & - & - \\
0.1 & - & 0.6991 & 0.7062 & 0.7002 & 0.7024 & 0.6997 & 0.7155 & 0.7073 & 0.7095 & 0.7008 & 0.7057 & 0.7062 & 0.6964 \\
0.2 & - & 0.7057 & 0.7089 & 0.6964 & 0.7067 & 0.7029 & 0.7127 & 0.6980 & 0.7095 & 0.7127 & 0.7089 & 0.7029 & 0.7089 \\
0.3 & - & 0.7018 & 0.7095 & 0.7013 & 0.7024 & 0.7057 & 0.7127 & 0.7051 & 0.7144 & 0.7062 & 0.7051 & 0.7095 & 0.7127 \\
0.4 & - & 0.7024 & 0.7067 & 0.7008 & 0.7095 & 0.6980 & 0.7013 & 0.7067 & 0.7138 & 0.7198 & 0.7100 & 0.7127 & 0.7078 \\
0.5 & - & 0.6997 & 0.7057 & 0.7122 & 0.7084 & 0.6997 & 0.7040 & 0.7111 & 0.7127 & 0.7100 & 0.7002 & 0.7078 & 0.7008 \\
0.6 & - & 0.6937 & 0.6986 & 0.7073 & 0.6970 & 0.7046 & 0.7067 & 0.7067 & 0.7127 & 0.7160 & 0.7084 & 0.7029 & 0.6915 \\
0.7 & - & 0.6980 & 0.7062 & 0.7013 & 0.6953 & 0.7084 & 0.7029 & 0.7008 & 0.7160 & 0.7024 & 0.7057 & 0.7002 & 0.6959 \\
0.8 & - & 0.7057 & 0.6921 & 0.7018 & 0.7024 & 0.6980 & 0.6991 & 0.7008 & 0.7073 & 0.7062 & 0.7018 & 0.7024 & 0.6991 \\
0.9 & - & 0.6877 & 0.6975 & 0.6942 & 0.6866 & 0.6877 & 0.6834 & 0.6861 & 0.6948 & 0.6910 & 0.6882 & 0.6882 & 0.6757 \\
\hline
\end{tabular}%
}
\end{table}

\begin{table}[htbp]
\caption{Final Winogrande evals with varying $r$ and $s$, for 600M models trained for 20k total steps. See Table~\ref{tab:results} for training details.}
\label{tab:winogrande}
\centering
\resizebox{\textwidth}{!}{%
\begin{tabular}{c|ccccccccccccc}
\hline
\diagbox{$r$}{$s$} & 1 & 2 & 3 & 4 & 5 & 6 & 7 & 8 & 9 & 10 & 11 & 12 & 13 \\
\hline
0.0 & 0.5257 & -- & -- & -- & -- & -- & -- & -- & -- & -- & -- & -- & -- \\
0.1 & -- & 0.5359 & 0.5446 & 0.5493 & 0.5438 & 0.5493 & 0.5533 & 0.5335 & 0.5478 & 0.5517 & 0.5604 & 0.5430 & 0.5422 \\
0.2 & -- & 0.5643 & 0.5280 & 0.5493 & 0.5675 & 0.5462 & 0.5478 & 0.5406 & 0.5359 & 0.5564 & 0.5485 & 0.5517 & 0.5399 \\
0.3 & -- & 0.5328 & 0.5320 & 0.5430 & 0.5509 & 0.5391 & 0.5620 & 0.5328 & 0.5320 & 0.5454 & 0.5501 & 0.5454 & 0.5596 \\
0.4 & -- & 0.5288 & 0.5422 & 0.5193 & 0.5509 & 0.5501 & 0.5667 & 0.5320 & 0.5383 & 0.5375 & 0.5422 & 0.5438 & 0.5422 \\
0.5 & -- & 0.5328 & 0.5493 & 0.5375 & 0.5375 & 0.5517 & 0.5604 & 0.5391 & 0.5501 & 0.5446 & 0.5399 & 0.5264 & 0.5754 \\
0.6 & -- & 0.5509 & 0.5391 & 0.5249 & 0.5501 & 0.5470 & 0.5509 & 0.5304 & 0.5454 & 0.5343 & 0.5580 & 0.5406 & 0.5596 \\
0.7 & -- & 0.5517 & 0.5549 & 0.5335 & 0.5438 & 0.5478 & 0.5509 & 0.5391 & 0.5335 & 0.5264 & 0.5406 & 0.5280 & 0.5422 \\
0.8 & -- & 0.5335 & 0.5399 & 0.5501 & 0.5359 & 0.5525 & 0.5383 & 0.5328 & 0.5304 & 0.5162 & 0.5296 & 0.5178 & 0.5241 \\
0.9 & -- & 0.5335 & 0.5414 & 0.5485 & 0.5335 & 0.5509 & 0.5406 & 0.5154 & 0.5296 & 0.5043 & 0.5162 & 0.5241 & 0.5107 \\
\hline
\end{tabular}%
}
\end{table}

\end{document}